\documentclass[final]{cvpr}

\usepackage{times}
\usepackage{epsfig}
\usepackage{graphicx}
\usepackage{amsmath}
\usepackage{amssymb}

\usepackage[dvipsnames]{xcolor}
\usepackage{algorithm}
\usepackage{subcaption}
\usepackage[noend]{algpseudocode}
\usepackage{ctable} % for \specialrule command
\usepackage[square,numbers,sort&compress]{natbib}
\usepackage[font=small,labelfont=bf]{caption}
\usepackage{color, colortbl}
\usepackage{listings}
\usepackage{mwe} % for placeholder images
\usepackage{makecell}
\usepackage{multirow}

\usepackage[scaled=0.85]{DejaVuSansMono}

\renewcommand{\algorithmicrequire}{\textbf{Input:}}
\renewcommand{\algorithmicensure}{\textbf{Output:}}

\newcommand{\cP}{{\cal P}}
\newcommand{\bp}{{\bf p}}

\newcommand{\by}{{\bf y}}
\newcommand{\cD}{{\cal D}}
\newcommand{\bS}{{\bf S}}
\newcommand{\bU}{{\bf U}}
\newcommand{\cR}{{\cal R}}
\newcommand{\br}{{\bf r}}
\newcommand{\bn}{{\bf n}}
\newcommand{\cS}{{\cal S}}
\newcommand{\cN}{{\cal N}}

\newcommand{\bphi}{{\boldsymbol \phi}}
\newcommand{\bpsi}{{\boldsymbol \psi}}
\newcommand{\bmu}{{\boldsymbol \mu}}
\newcommand{\bxi}{{\boldsymbol \xi}}

\newcommand{\NAME}{WyPR}

\definecolor{chair}{RGB}{66,103,178}
\definecolor{table}{RGB}{232,74,39}
\definecolor{sofa}{RGB}{140, 86, 75}
\definecolor{bookshelf}{RGB}{148, 103, 189}
\definecolor{highlightRowColor}{rgb}{0.95, 0.95, 1}
\definecolor{darkgreen}{RGB}{56,137,35}
\definecolor{brown}{RGB}{132,60,12}
\definecolor{darkblue}{RGB}{0,32,96}

\newcolumntype{H}{>{\setbox0=\hbox\bgroup}c<{\egroup}@{}}  % to hide a column

\usepackage[pagebackref=true,breaklinks=true,bookmarks=false,colorlinks,citecolor=ForestGreen]{hyperref}
\usepackage{cleveref}

\crefname{section}{\S}{\S\S}
\crefname{subsection}{\S}{\S\S}
\crefformat{table}{Tab.~#2#1#3}
\crefformat{figure}{Fig.~#2#1#3}
\crefformat{equation}{Eq.~(#2#1#3)}
\crefformat{algorithm}{Alg.~#2#1#3}
\crefformat{appendix}{Appendix~#2#1#3}

\def\cvprPaperID{1136} % *** Enter the CVPR Paper ID here
\def\confYear{CVPR 2021}

\begin{document}

%%%%%%%%% TITLE
\title{3D Spatial Recognition without Spatially Labeled 3D}

\author{
Zhongzheng Ren$^{1,2\footnotemark{}}$ \qquad Ishan Misra$^{1}$ \qquad Alexander G. Schwing$^{2}$ \qquad Rohit Girdhar$^{1}$ \\
{$^{1}$Facebook AI Research \qquad $^{2}$University of Illinois at Urbana-Champaign} \\
{\normalsize \url{https://facebookresearch.github.io/WyPR}}
}

% \maketitle

\twocolumn[{
\renewcommand\twocolumn[1][]{#1}
\maketitle
\begin{center}
\includegraphics[width=\textwidth]{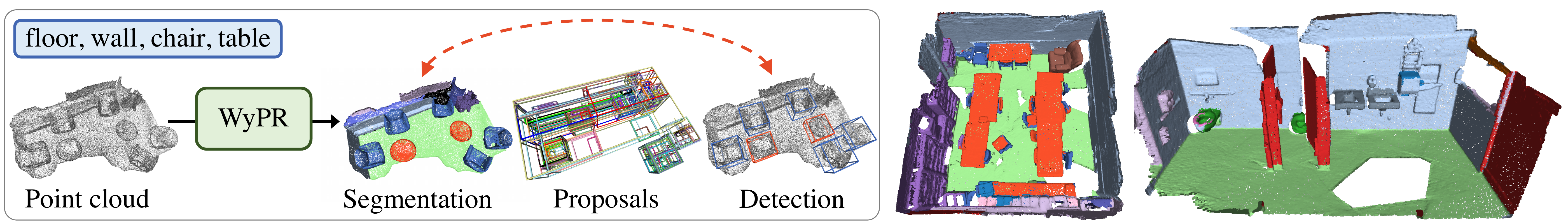}
\captionof{figure}{(Left) Our framework, \NAME{}, jointly learns semantic segmentation  and object detection for point cloud data from only scene-level class tags. We find that encouraging consistency between the two tasks is key. (Right) Sample segmentation results from ScanNet val set, without seeing any point-level labels during training. Please refer to~\cref{sec:analysis} and~\cref{supp:vis} for more analysis and visualizations.
}
\label{fig:teaser}
\end{center}%
}]

%%%%%%%%% ABSTRACT
\begin{abstract}

We introduce \NAME{}, a \textbf{W}eakl\textbf{y}-supervised framework for \textbf{P}oint cloud \textbf{R}ecognition,
requiring only scene-level class tags as supervision.
\NAME{} jointly addresses three core 3D recognition tasks: point-level semantic segmentation, 3D proposal generation, and 3D object detection, coupling their predictions through self and cross-task consistency losses. We show that in conjunction with standard multiple-instance learning objectives, \NAME{} can detect and segment objects in point cloud data without access to any spatial labels at training time. We demonstrate its efficacy using the ScanNet and S3DIS datasets,  outperforming prior state of the art on weakly-supervised segmentation by more than 6\% mIoU. In addition, we set up the first benchmark for weakly-supervised 3D object detection on both datasets, where \NAME{} outperforms standard approaches and establishes strong baselines for future work.

\end{abstract}

\footnotetext{$^*$Work partly done during an internship at Facebook AI Research.}

%%%%%%%%% BODY TEXT
%!TEX root = ../main.tex
\vspace{-1em}
\section{Introduction}
\label{sec:intro}

Recognition (\ie, segmentation and detection) of 3D objects is a key step towards scene understanding. With the recent development of consumer-level depth sensors (\eg, LiDAR~\cite{franklin2020ipad,stein2020iphone}) and the advances of computer vision algorithms, 3D data collection has become more convenient and inexpensive. However, existing 3D recognition systems often fail to scale as they rely on strong supervision, such as point level semantic labels or 3D bounding boxes~\cite{qi2017pointnet++, qi2019votenet, choy20194d}, which are time consuming to obtain. For example, while the popular large-scale indoor 3D dataset ScanNet~\cite{dai2017scannet} was collected by only 20 people, the annotation effort involved more than 500 annotators spending nearly 22.3 minutes per scan. Furthermore, due to the high annotation cost, existing 3D object detection datasets have limited themselves to a small number of object classes. This time consuming labeling process is a major bottleneck preventing the community from scaling  3D recognition.

Motivated by this observation, we study 3D weakly-supervised learning with only scene-level class tags available as supervision to train semantic segmentation and object detection models. Scene-level tags are very efficient to annotate, taking only a second or less for each object in the scene~\cite{ren-eccv2020}. Hence, methods that rely on such supervision can be scaled more easily  than those that rely on box-level supervision. 

\begin{table*}[t]
\centering
\small
\begin{tabular}{c|cccc|cc|c}
\specialrule{.15em}{.05em}{.05em}
Methods & ~\cite{tang2019transferable}  & \cite{wilson20203d} & \cite{xu2020weakly} & \cite{wang2019towards} & \cite{wei2020multi} & \cite{qin2020weakly}  & \NAME{} \\
\hline
Weak labels & 2D boxes & 2D inst seg & sparse label & 2D sem seg & region \& scene tags & scene tags & scene tags  \\
Tasks & det & det & seg & seg & seg & det & det + seg \\
Dataset & indoor  & outdoor & indoor \& objects & indoor & indoor & outdoor & indoor\\
\specialrule{.15em}{.05em}{.05em}
\end{tabular} 
\vspace{-0.2cm}
\caption{{\bf Summary of closely related work in weakly-supervised 3D recognition.} Compared to prior work, our proposed method (\NAME) uses the readily available scene tags, and jointly learns detection and segmentation in the more challenging indoor room setting.
}
\label{tab:related}
\vspace{-0.2cm}
\end{table*}

For this we develop the novel weakly-supervised framework called \NAME{},  shown in~\cref{fig:teaser}. Using just scene level tags, it jointly learns both segmentation of point cloud and detection of 3D boxes. Why should \emph{joint} learning of segmentation and detection perform better than independently learning the two tasks? First, since these two tasks are related, joint training is mutually beneficial for representation learning. Second, these tasks naturally constrain each other, leading to effective self-supervised objectives that further improve performance. For example, the semantic labels of points within a bounding box should be consistent, and vice versa. Lastly, directly learning to regress to dimensions of 3D bounding boxes, as common in supervised approaches~\cite{qi2019votenet,Shi_2019_CVPR,qi2020imvotenet}, is extremely challenging using weak labels. Learning weakly-supervised segmentation first permits a two-stage detection framework, where object proposals are generated bottom-up conditioned on segmentation prediction and further classified using a weakly-supervised detection algorithm.

To achieve this, \NAME{} operates on point cloud data of complex indoor scenes and combines a weakly-supervised semantic segmentation stage (\cref{sec:seg}) with a weakly-supervised object detection stage (\cref{sec:det}).
The latter takes as input the geometric representation of the input scene and a set of computed 3D proposals from {\bf GSS}, our novel {\bf G}eometric {\bf S}elective {\bf S}earch algorithm (\cref{sec:weak_prop}). GSS uses local geometric structures (\eg, planes) and the previously computed segmentation, for bottom-up proposal generation. Due to the uninformative nature of weak labels, weakly-supervised frameworks often suffer from noisy prediction and high variance. We address this by encouraging both cross-task and cross-transformation consistency through self-supervised objectives. We evaluate \NAME{} on standard 3D datasets, \ie, ScanNet and S3DIS (\cref{sec:exp}), improving over prior work on weakly-supervised 3D segmentation by more than $6$\% mIoU, and establishing  new benchmarks and strong baselines for weakly-supervised 3D detection.

Our contributions are as follows:~1) a novel point cloud framework to jointly learn weakly-supervised semantic segmentation and object detection, which significantly outperforms single task baselines;~2) an unsupervised 3D proposal generation algorithm, geometric selective search (GSS), for point cloud data;~and 3) state-of-the-art results on weakly-supervised semantic segmentation, and benchmarks on weakly-supervised proposal generation and object detection.

\begin{figure*}[t]
\vspace{-1em}
\centering
\includegraphics[width=\linewidth]{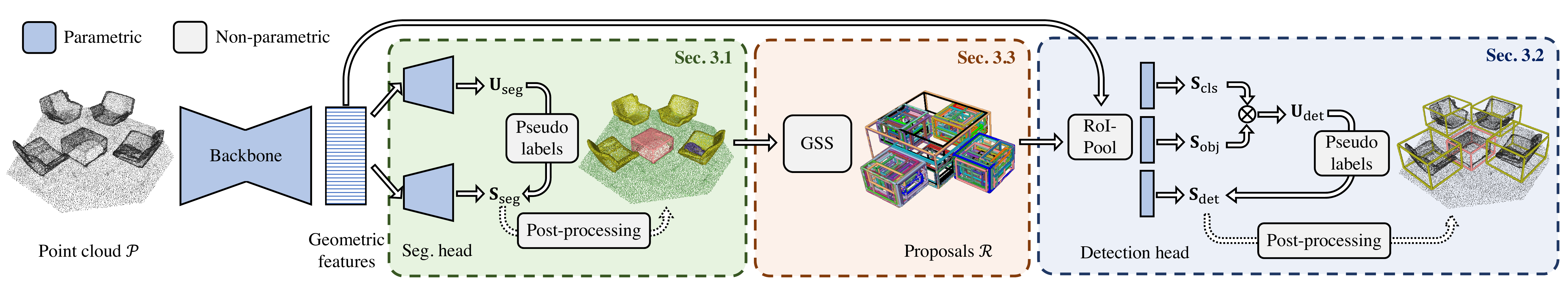}
\caption{{\bf Approach Overview.} A backbone network extracts geometric features which are used by the segmentation head to compute a point-level segmentation map. The segmentation map is passed into the 3D proposal generation module GSS, and the resulting proposals along with original features are used to detect 3D object instances. Through a series of self and cross-task consistency losses along with multiple-instance learning objectives, \NAME{} is trained end-to-end using only scene-level tags as supervision.}
\label{fig:archi}
\vspace{-0.3cm}
\end{figure*}

%!TEX root = ../main.tex
\section{Related work}

\par\noindent\textbf{3D datasets.}
Semantically labeled 3D data can be broadly classified into indoor~\cite{armeni2017joint,song2015sun,dai2017scannet,Matterport3D} and outdoor~\cite{chang2019argoverse,sun2020scalability,geiger2013vision,choi2016large} settings. ScanNet~\cite{dai2017scannet}, a popular 3D detection and segmentation dataset, contains 20 classes labeled in 1500 scenes. While this dataset is large, it is small in comparison to 2D datasets, which reach tens of millions of images~\cite{kuznetsova2018open} and thousands of instance labels~\cite{gupta2019lvis}. While the popularity of advanced 3D sensors~\cite{stein2020iphone,franklin2020ipad} could lead to a similar growth in 3D data, annotating that data would still be extremely time consuming. This underscores the need to develop weakly-supervised techniques for 3D recognition.

\par\noindent\textbf{3D representations.}
3D data is often represented via a point cloud, and processed using one of two main backbone architectures. The first~\cite{girdhar16learning,riegler2017octnet,choy20194d,SparseConvNet} projects points to intermediate volumetric grids, and then processes them using convolutional nets. These methods are efficient but suffer from information loss due to the discretization into voxels. The second operates directly on points~\cite{qi2017pointnet++,qi2017pointnet,Wang2018DynamicGC,thomas2019KPConv}, processing them in parallel either using a pointwise MLP~\cite{qi2017pointnet,qi2017pointnet++}, graph convolution~\cite{Wang2018DynamicGC}, or point convolution~\cite{thomas2019KPConv}. Our method is compatible with either backbone architecture. We adopt PointNet++~\cite{qi2017pointnet++} for experimentation. 

\par\noindent\textbf{3D tasks.}
Semantic segmentation~\cite{dai2017scannet, shapenet2015,armeni2017joint}, object detection~\cite{qi2019votenet, qi2018frustum, Shi_2019_CVPR, song2016deep}, and classification~\cite{wu20153d} are the standard recognition tasks defined on 3D data. For segmentation, the two most common tasks are point-level object parts segmentation~\cite{shapenet2015} and scene object segmentation~\cite{dai2017scannet,armeni2017joint}, the latter of which we address in this work using weak supervision. For 3D object detection, standard techniques leverage either only a point cloud~\cite{qi2019votenet,Shi_2019_CVPR,zhang2020h3dnet}, or a point cloud together with the corresponding multi-view RGB images~\cite{hou20193d,qi2020imvotenet,qi2018frustum}. Unlike 2D where offline proposal generation methods~\cite{eb,ss} are widely studied and generalize well to unseen datasets, 3D proposals generated from a point cloud are often trained in a supervised manner~\cite{ku2018joint, qi2019votenet, Shi_2019_CVPR} and overfit to the training set. We propose an unsupervised 3D proposal generation algorithm GSS, which we further improve using weak supervision.

\par\noindent\textbf{Weakly-supervised learning.}
Weak labels in the form of image-level class tags are widely studied in 2D tasks such as image localization~\cite{Zhou_2016_CVPR,XuCVPR2014}, semantic segmentation~\cite{wei2018revisiting, pathakICCV15ccnn}, and object detection~\cite{bilen2016weakly, tang2017multiple, ren-cvpr2020}. Prior work mostly formulates weakly-supervised learning as a multiple instance learning problem, where the target tasks are learned implicitly in a multi-label classification framework. Pipelined~\cite{wei2017object,krishna-cvpr2016} or end-to-end self-training modules~\cite{tang2017multiple,ren-cvpr2020} have also been demonstrated to be beneficial.

\par\noindent\textbf{Weakly-supervised learning in 3D.}
Compared to its 2D counterpart, weakly-supervised learning for 3D tasks is relatively unexplored. We  summarize all relevant prior work in~\cref{tab:related}. For semantic segmentation, Wang~\etal~\cite{wang2019towards} leverage 2D segmentation as weak labels, Xu~\etal~\cite{xu2020weakly} use a sparsely labeled point cloud, and Wei~\etal~\cite{wei2020multi} utilize both area and scene-level class tags during training. For object detection, recent work uses small sets of labeled 3D data~\cite{meng2020weakly,tang2019transferable,Zhao_2020_CVPR}, 2D instance segmentation~\cite{wilson20203d}, and click annotation~\cite{meng2020weakly} as supervision. However, obtaining these labels is still time consuming. A closely related concurrent work~\cite{qin2020weakly} focuses on autonomous driving, building upon a small number of relatively easy objects (\eg, car, pedestrian) while still using image data. In contrast, we focus on complex indoor scenes, exclusively relying on the 3D point cloud, \ie, no images are required. 

\noindent\textbf{Multi-task learning.}
Multi-task learning~\cite{mtl} has been widely studied for various vision tasks~\cite{eigen-iccv2015, ren-cvpr2018, ubernet, cross-stitch}. It is of particular importance for weakly-supervised~\cite{ren-eccv2020,li2019weakly} or self-supervised 2D object detection~\cite{ren-cvpr2018, doersch2017multitask} as multi-tasking provides mutual regularization and hence better representation learning. For detection and segmentation, prior work has studied joint training with 2D data~\cite{li2019weakly} or supervised 3D data~\cite{unal2020improving}. In this paper, we develop a novel framework for learning both tasks under weak supervision.

%!TEX root = ../main.tex
\section{\NAME}
\label{sec:app}

Our goal is to use weak supervision in the form of scene-level tags and learn a joint 3D segmentation and detection model, which we refer to as \textbf{\NAME}. Specifically, we assume availability of data $\cD=\{(\cP, \by)\}$ of point cloud $\cP$ and corresponding scene-level tags $\by\in\{0,1\}^C$, which indicate absence or presence of the $C$ object classes. $\cP$ is a set of six-dimensional points $\bp\in\cP$, represented by their 3D location and RGB-color. Note, $\by$ only indicates  existence of objects in the  scene and does not contain any information about per-point semantic labels or object locations.

\par \noindent \textbf{Approach overview.}~\cref{fig:archi} provides an overview of our model which consists of three parametric modules: a backbone network, followed by a segmentation and a detection head. We first extract geometric features from the input point cloud using the backbone network. Specifically, we use the variant of PointNet++~\cite{qi2017pointnet++} following VoteNet~\cite{qi2019votenet}, which is an encoder-decoder network with skip connections. The features are then fed into the segmentation and detection modules. The segmentation module assigns each point from the input point cloud $\cP$ to one of $C$ classes. We use this segmentation output to generate 3D region proposals $\cR$ that are likely to contain objects in the scene. Finally, the detection module classifies each proposal into either one of $C$ classes or background (not an object) class, using the backbone features corresponding to that proposal.

\par \noindent \textbf{Notation.} We denote the output of the segmentation module as $\bS_\text{seg}\in\mathbb{R}^{|\mathcal{P}|\times C}$, where the rows represent the score logits over the $C$ classes for all points $\cP$. The detection module produces a score matrix $\bS_\text{det}\in\mathbb{R}^{|\cR|\times (C+1)}$ over the $C$ classes and background for all 3D proposals $\cR$. For readability, we also use $\bp, \br$ as indices into $\bS_\text{seg}, \bS_\text{det}$ in the following sections.

\subsection{Weakly-supervised 3D semantic segmentation}
\label{sec:seg}

The segmentation module consists of two identical heads that independently process the backbone features using a series of unit PointNet~\cite{qi2017pointnet} and nearest neighbor upsampling layers ({\color{darkgreen} \cref{fig:archi} green region}). The output from these heads are two score matrices $\bU_\text{seg},\bS_\text{seg} \in \mathbb{R}^{|\mathcal{P}|\times C}$ respectively, containing logits over $C$ object classes for all  points $\bp\in\cP$. The parameters of the backbone and the segmentation module are optimized to minimize a composed loss
\begin{align}\label{eqn:loss_seg}
 \mathcal{L}_\text{seg} = \mathcal{L}_\text{seg}^{\textsc{MIL}} + \mathcal{L}_\text{seg}^{\textsc{SELF}} + \mathcal{L}_\text{seg}^{\textsc{CST}} + \mathcal{L}_{\text{d}\rightarrow \text{s}} + \mathcal{L}_\text{smooth} ,
\end{align}
where $\mathcal{L}_\text{seg}^{\textsc{MIL}}$ denotes a multiple-instance learning (MIL) loss, $\mathcal{L}_\text{seg}^{\textsc{SELF}}$ denotes a self-training loss, $\mathcal{L}_\text{seg}^{\textsc{CST}}$ and $\mathcal{L}_{\text{d}\rightarrow \text{s}}$ represent consistency loss across geometric transformations and tasks respectively, and $\mathcal{L}_\text{smooth}$ is a smoothness regularization loss. We describe the individual loss terms next.

\par \noindent \textbf{MIL loss.} The multiple-instance learning loss~\cite{wei2017object,wei2018revisiting} encourages to learn the per-point semantic segmentation logits without access to point-level supervision. We first convert the per-point logits $\bU_\text{seg}$ into a scene-level prediction $\bphi$ via average pooling and a sigmoid normalization
\begin{align}\label{eqn:avg-pool-seg}
\bphi[c] = \text{sigmoid}\left( \frac{1}{|{\mathcal{P}}|} \sum_{\bp\in\mathcal{P}} {\bf U}_\text{seg} [\bp, c] \right).
\end{align}
The scene-level prediction $\bphi$ is then supervised using the scene-level tags $\bf y$ using the binary cross-entropy loss
\begin{align}\label{eqn:mil-seg}
\mathcal{L}_\text{seg}^{\textsc{MIL}} \!=\!  -\!\sum_{c=1}^{C} {\bf y}[c] \log \bphi[c]   \!-\! (1\!-\!{\bf y}[c] )\log  (1\!-\! \bphi[c]).
\end{align}

\begin{algorithm}[t]
\footnotesize
\caption{Segmentation pseudo label generation}
\begin{algorithmic}[1]
\Require{class label $\by$, segmentation logits $\bU_\text{seg}$, threshold $p_1$}
\Ensure{pseudo label $\hat{\bf Y}_\text{seg}$}
\State $ \hat{\bf Y}_\text{seg} = \mathbf{0}$  \algorithmiccomment{initialize to zero matrix}
\For{each point $\bp \in \cP$}
\State 	$ c = \text{argmax}(\by \odot \bU_\text{seg}[\bp,:])$ \algorithmiccomment{element-wise product}
\State $\hat{\bf Y}_\text{seg}\left[\bp, c\right] = 1 $
\EndFor
\For{ground-truth class $c$ where $\by[c]=1$}
\State $\cP'[c] \leftarrow $  lowest $p_1\text{-th}$ percentile of $\hat{\bf Y}_\text{seg}[:, c]$
\State $\hat{\bf Y}_\text{seg}\left[\bp, c\right] = 0 \;\forall \bp \in \cP'[c]$  \algorithmiccomment{ignore points with low score}
\EndFor
\end{algorithmic}
\label{alg:pseudo-seg}
\end{algorithm}

\par \noindent \textbf{Self-training loss.}
Inspired by the success of self-training in weakly-supervised detection~\cite{tang2017multiple,wan2019c,ren-cvpr2020}, we further incorporate a self-training loss. The previously computed segmentation logits $\bU_\text{seg}$ are used to supervise the final segmentation logits $\bS_\text{seg}$ via a cross-entropy loss
\begin{align}\label{eqn:self-seg}
\mathcal{L}_\text{seg}^{\textsc{SELF}} = -\frac{1}{|\mathcal{P}|} \sum_{\bp\in \mathcal{P}}\sum_{c=1}^{C} {\bf \hat{Y}}_\text{seg}[\bp, c] \log \bpsi[\bp, c],
\end{align}
where $\bpsi[\bp,c]=\text{softmax}(\bS_\text{seg}[\bp, c])$ denotes the probability of point $\bp$ belonging to class $c$, and ${\bf \hat{Y}}_\text{seg}[\bp,c] \in \{0,1\}$ is the point-level pseudo class label inferred from score matrix $\bU_\text{seg}$.
We detail the process of computing the pseudo label in~\cref{alg:pseudo-seg}. Intuitively, the algorithm ignores noisy predictions in $\bU_\text{seg}$, leading to robust self-supervision for $\bS_\text{seg}$.

\par \noindent \textbf{Cross-transformation consistency loss.} In addition, we use $\mathcal{L}_\text{seg}^{\textsc{CST}}$ to encourage that the segmentation predictions are consistent across data augmentations $\mathcal{T}$. We obtain an augmented point cloud $\tilde{\mathcal{P}}=\mathcal{T} (\mathcal{P})$ by changing the original scene $\mathcal{P}$ via standard augmentations (see~\cref{sec:implement} and~\cref{supp:aug} for details). We predict the semantic segmentation $\tilde{\bS}_\text{seg}$ on this transformed point cloud. The consistency loss is then formulated as 
\begin{align}\label{eqn:consis-seg}
\mathcal{L}_\text{seg}^{\textsc{CST}} \!=\! \frac{1}{|\mathcal{P}\cap \mathcal{\tilde{P}}|}\!\sum_{\bp\in\mathcal{P}\cap\mathcal{\tilde{P}}} D_{\textsc{KL}} \left( \bpsi[\bp,\cdot] \left|\right| \tilde{\bpsi}[\bp,\cdot] \right),
\end{align}
where $\bpsi[\bp,c]\!=\!\text{softmax}(\bS_\text{seg}[\bp, c])$ and $\tilde{\bpsi}[\bp,c]\!=\!\text{softmax}(\tilde{\bS}_\text{seg}[\bp, c])$ are the probabilities of the point $\bp$ belonging to class $c$, and $D_\textsc{KL}$ is the KL divergence over $C$ classes for points that are common across the transformation. This loss encourages  the probability distributions for semantic segmentation of corresponding points within the point cloud $\cP$ and $\tilde{\cP}$ to match. 

\par \noindent \textbf{Cross-task consistency loss.}
We further employ a cross-task regularization term $\mathcal{L}_{\text{d}\rightarrow \text{s}}$. It uses the detection results to refine the segmentation prediction. Intuitively, all  points within a confident bounding box prediction should have the same semantic label.  Assume we have access to a  set of confident bounding boxes $\br\in\mathcal{R^\ast}$ and their corresponding predicted score matrix $\bS_\text{det}\in\mathbb{R}^{|\mathcal{R}^\ast|\times (C+1)}$. Using this information, we encourage consistency via a cross entropy loss on the point-level predictions, with the box-level prediction as a soft target
\begin{align}\label{eqn:d2s}
\mathcal{L}_{\text{d}\rightarrow \text{s}} \!=\! - \frac{1}{|\mathcal{R^\ast}|} \!\sum_{\br\in\mathcal{R^\ast}}\!  \frac{1}{|\mathcal{P}^r|} \sum_{\bp\in \mathcal{P}^\br}  \sum_{c=1}^{C}  \bxi[\br,c] \log \bpsi[\bp, c],
\end{align}
where $\bpsi[\bp, c]$ is the point probability from \cref{eqn:self-seg}, $\bxi[\br,c]=\text{softmax}(\bS_\text{det}[\br, c])$ denotes the probability of proposal $\br$ belonging to object class $c$, and $\mathcal{P}^\br$ denotes the set of points within proposal $\br$.
In practice, the confident bounding boxes $\mathcal{R^\ast}$ are obtained from~\cref{alg:pseudo-det}, discussed later in~\cref{sec:det}.

\par \noindent \textbf{Smoothness regularization.}
Finally, we compute $\mathcal{L}_\text{smooth}$ to encourage local smoothness. We first detect a set of planes  $\mathcal{G}$ from input point cloud $\mathcal{P}$ using an unsupervised off-the-shelf shape detection algorithm~\cite{lafarge2012creating} detailed in~\cref{supp:shape}.  We then compute
\begin{align}\label{eqn:smo}
\mathcal{L}_\text{smooth} = -  \sum_{i=1}^{|\mathcal{G}|} \frac{1}{|\mathcal{G}[i]|} \sum_{\bp\in \mathcal{G}[i]}  \sum_{c=1}^{C}  \bar{\bpsi}[c] \log \bpsi[\bp, c],
\end{align}
where $\bar{\bpsi}[c] =\frac{\sum_{\bp\in\mathcal{G}[i]}\bpsi[\bp,c]}{|\mathcal{G}[i]|}$ is the mean probability of all the points which lie inside plane $\mathcal{G}[i]$ for class $c$.

\begin{figure*}[t]
\centering
\includegraphics[width=\textwidth]{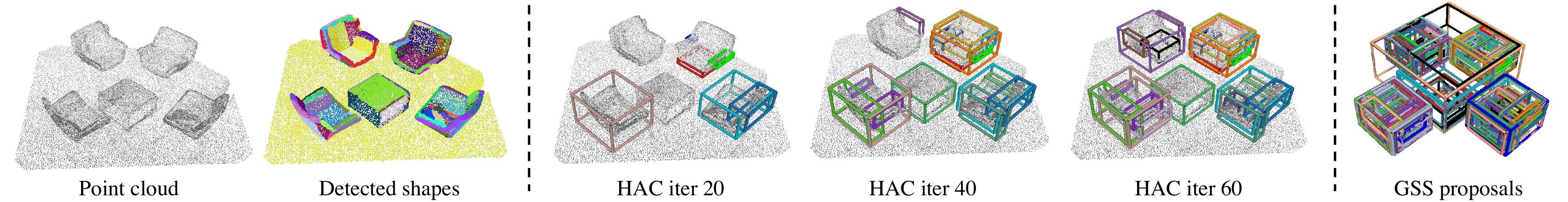}
\caption{{\bf Geometric Selective Search (GSS).} Our algorithm takes as input the point cloud and detected planes (left column). It then hierarchically groups the neighboring planes into sub-regions  and generates 3D proposals for the combined regions (middle column). We run the algorithm multiple times with different grouping criteria to encourage high recall of final output proposals (right column).}
\label{fig:prop}
\vspace{-0.1cm}
\end{figure*}

\begin{algorithm}[t]
\footnotesize
\caption{Detection pseudo label generation}
\begin{algorithmic}[1]
\Require{class label $\by$, detection logits $\bU_\text{det}$, proposals $\mathcal{R}$, threshold $\tau, p_2$}
\Ensure{pseudo label $\hat{\bf Y}_\text{det}$}
\For{ground-truth class $c$ where $\by[c]=1$}
\State 	$ \hat{\bf Y}_\text{det} = \mathbf{0}$  \algorithmiccomment{initialize to zero matrix}
\State $\mathcal{R}'[c] \leftarrow $  top $p_2\text{-th}$ percentile of $\bU_\text{det}[:, c]$  \algorithmiccomment{$\mathcal{R}'[c]$ is descending}
\State  $ \mathcal{R^*}[c] \leftarrow \br'_1 $   \algorithmiccomment{save 1st RoI (top-scoring) $ \br'_1 \in \mathcal{R}'[c]$}
\For{$i\in\{2,\cdots,|\mathcal{R}'[c]|\}$}   \algorithmiccomment{~start from the 2nd highest}
\State $ \mathcal{R}^*[c] \leftarrow  \br'_i$ \textbf{if}~ $\text{IoU}(\br'_i, \hat \br) < \tau \;\forall \hat \br \in \mathcal{R}^*[c]$
\EndFor
\State $\hat{\bf Y}_\text{det}[\br,c]=1 \;\forall \br \in \mathcal{R}^*[c]$
\EndFor
\end{algorithmic}
\label{alg:pseudo-det}
\end{algorithm}

\subsection{Weakly-supervised 3D object detection}
\label{sec:det}
Our object detection module assumes access to a set of 3D region proposals $\mathcal{R}$ (discussed in~\cref{sec:weak_prop}) and uses the backbone features to classify the proposals into one of the $C$ object classes or background ({\color{darkblue}\cref{fig:archi} blue region}).
Each region of interest (RoI) $\br\in\mathbb{R}^6$ is represented by a six-dimensional vector denoting  its center location and its width, height and length.
We extract RoI features by averaging the backbone features of all the points within each proposal.
Inspired by prior 2D literature~\cite{bilen2016weakly},
we use three separate linear layers to extract classification logits $\bS_\text{cls}\in \mathbb{R}^{|\mathcal{R}|\times (C+1)}$, objectness logits $\bS_\text{obj}\in \mathbb{R}^{|\mathcal{R}|\times (C+1)}$, and final detection logits $\bS_\text{det} \in \mathbb{R}^{|\mathcal{R}|\times (C+1)}$ from the RoI features.
As in~\cite{bilen2016weakly}, we normalize $\bS_\text{cls}$ using a softmax function over rows to obtain the probability over object classes for each proposal. Similarly, we normalize $\bS_\text{obj}$ over columns to obtain a probability over proposals for each class.
Intuitively, $\bS_\text{cls}[\br,c]$ represents the probability of region $\br$ being classified as  class $c$, and $\bS_\text{obj}[\br,c]$ is the probability of detecting region $\br$ for  class $c$.
We aggregate the evidence from both matrices via element-wise multiplication to obtain the score matrix $\bU_\text{det} =\bS_\text{cls}\odot\bS_\text{obj}$.
Similar to the self-training discussed earlier for segmentation, we infer pseudo-labels from $\bU_\text{det}$ to supervise the final detection logits $\bS_\text{det}$. We learn the backbone and the detection module using the loss
\begin{align}\label{eqn:loss_det}
 \mathcal{L}_\text{det} = \mathcal{L}_\text{det}^{\textsc{MIL}} +  \mathcal{L}_\text{det}^{\textsc{SELF}} + \mathcal{L}_\text{det}^{\textsc{CST}},
\end{align}
where $\mathcal{L}_\text{det}^{\textsc{MIL}}$ is a MIL objective for detection, $\mathcal{L}_\text{det}^{\textsc{SELF}}$ is a self-training loss, and $\mathcal{L}_\text{det}^{\textsc{CST}}$ is the cross-transformation consistency loss. All the terms are described next.

\par \noindent \textbf{MIL loss.} Similar to the segmentation head, the multiple instance learning (MIL) loss for detection is 
\begin{align}\label{eqn:mil-det}
\mathcal{L}_\text{det}^{\textsc{MIL}} \!=\! -\sum_{c=1}^{C+1} \by[c] \log \bmu[c] \!-\! (1\!-\!\by[c])\log (1\!-\!\bmu[c]) ,
\end{align}
where $\bmu[c] = \sum_{\br\in \mathcal{R}} \bU_\text{det} [\br, c]$  is the row-sum of the score matrix $\bU_\text{det}$ for class $c$.
This sum-pooling operation aggregates RoI scores into a scene-level score vector $\bmu$, which is used for multi-label scene classification.

\par \noindent \textbf{Self-training loss.} As done before for segmentation,
we incorporate a self-training loss for detection as well. The final detection logits $\bS_\text{det}$ are supervised by $\bU_\text{det}$ via
\begin{align}\label{eqn:self-det}
\mathcal{L}_\text{det}^{\textsc{SELF}} = -\frac{1}{|\mathcal{R}|} \sum_{\br\in \mathcal{R} }\sum_{c=1}^{C+1} {\bf \hat{Y}}_\text{det}[\br, c] \log \bxi[\br, c],
\end{align}
where $\bxi[\br, c]=\text{softmax}(\bS_\text{det} [\br, c])$ denotes the probability of proposal $\br$ belonging to object class $c$, and  ${\bf \hat{Y}}_\text{det}[\br,c] \in \{0,1\}$ is the RoI pseudo class label inferred from score matrix $\bU_\text{det}$. The pseudo label ${\bf \hat{Y}}_\text{det}$ is computed using~\cref{alg:pseudo-det}. Conceptually, this algorithm selects a set of confident yet diverse predictions as the pseudo labels for self-training.

\par \noindent \textbf{Cross-transformation consistency loss.} Following the consistency loss for semantic segmentation (\cref{eqn:consis-seg}), we  encourage  detection predictions to be consistent under  transformation $\mathcal{T}$ via
\begin{align}\label{eqn:consis-det}
\mathcal{L}_\text{det}^{\textsc{CST}} = \frac{1}{|{\cal R}|}\sum_{\br\in\mathcal{R}} D_{\textsc{KL}} \left( \bxi \big[\br,\cdot] \;\left|\right|\; \tilde{\bxi}[\mathcal{T} (\br),\cdot \big] \right),
\end{align}
where $\bxi[\br,c]$ refers to the RoI probability  introduced in~\cref{eqn:self-det}, $\tilde{\bxi}[\mathcal{T}(\br),c]$ denotes the RoI probability obtained from the transformed input $\tilde{\mathcal{P}}=\mathcal{T} (\mathcal{P})$ and proposal $\mathcal{T}(\br)$ via the same backbone  and detection module. %\as{symmetrized KL here?}

\subsection{Geometric Selective Search (GSS)}
\label{sec:weak_prop}
The detection module uses a proposal set ${\cal R}$ as input.
In weakly-supervised learning, proposals are necessary because it is not possible to mimic supervised methods that directly predict 3D bounding box parameters (\eg, size and location).
The key observation which inspires our novel 3D proposal generation algorithm is that most indoor  objects are rigid and mainly consist of basic geometric structures (\eg, planes, cylinders, spheres). We thus devise a bottom-up solution termed Geometric Selective Search (GSS), first detecting basic geometric shapes which are then grouped hierarchically to form 3D proposals ({\color{brown}\cref{fig:archi} brown region}).

Given an input point cloud with unoriented normals, we adopt a region-growing-based method~\cite{lafarge2012creating,cgal:ovja-pssd-20b} for detecting primitive shapes (\eg, planes) as shown in~\cref{fig:prop} left.  We choose region-growing over the popular RANSAC-based methods~\cite{schnabel2007efficient}  because 1) it is deterministic; 2) it performs better in the presence of large scenes with fine-grained details. We then apply hierarchical agglomerative clustering (HAC) to iteratively group the detected shapes into sub-regions. In each HAC iteration, we compute the similarity score $s$ between all spatially overlapping sub-regions and group the two most similar regions. We iterate until no neighbors can be found or only one region is left.
Every time we generate a new region, we also compute the axis-aligned bounding boxes of the new region and add it into the proposal pool. 
We illustrate the process of growing the proposal pool during HAC in~\cref{fig:prop} (middle columns). 

In order to pick which two regions $\bn_i, \bn_j$ to group, HAC uses a similarity score
\begin{align} \label{eq:prop-sim}
s(\bn_i, \bn_j) =  w_1 s_\text{size} \!+\!  w_2 s_\text{volume} \!+\! w_3 s_\text{fill} \!+\! w_4 s_\text{seg},
\end{align}
where $w_i \in \{0,1\} ~~\forall i\in \{1,\cdots,4\}$ are binary indicators. $s_\text{size}$ and $s_\text{volume}\in[0,1]$ measure size and volume compatibility and encourage small regions to merge early; $s_\text{fill}\in[0,1]$ measures how well two regions are aligned.
Besides similarities of low-level cues, we also measure high-level semantic similarities by incorporating segmentation similarity $s_\text{seg}\in[0,1]$.
This score is the histogram intersection of the normalized $C$-dimensional class histogram of two regions' points. The class labels of these points are computed from $\bS_\text{seg}$ using the inference procedure described in~\cref{sec:implement}.
Please see~\cref{sec:proposal} for the exact formulation of the above metrics.
During training, as the segmentation module improves, $s_\text{seg}$ increasingly prefers grouping regions which correspond to the same object. A similar idea to compute proposals from segmentations has also been widely adopted in the 2D case~\cite{ss,mcg}. In practice, we find that multiple runs of HAC with different $w_i$ values, results in a more diverse set of proposals as each run uses a different weighted similarity measure. We provide the values of $w_i$ for different runs in~\cref{sec:proposal}.

GSS can be made \textbf{completely unsupervised} by removing the segmentation term $s_\text{seg}$ from~\cref{eq:prop-sim}. This variant is also valuable as the proposals can be pre-computed offline and are of decent quality (verified in~\cref{sec:exp}). These proposals
are independent of any specific supervision and can benefit various downstream unsupervised or weakly-supervised 3D recognition tasks, akin to Selective Search~\cite{ss} or Edge Boxes~\cite{eb} in 2D. This is distinct from existing 3D proposal techniques that either use 2D image cues~\cite{qin2020weakly} or full bounding box supervision~\cite{qi2020imvotenet,qi2019votenet}.
%!TEX root = ../main.tex
\section{Experiments}
\label{sec:exp}
\label{sec:implement}

We empirically evaluate \NAME{} on two standard 3D benchmarks. We first provide the key implementation details (more details in~\cref{supp:imp}) and describe the baseline methods we compare to (\cref{sec:baseline}). We then present the quantitative results (\cref{sec:quant} and~\ref{sec:quant-s3dis}), ablate our design choices and present qualitative results (\cref{sec:qualitative}).

\par \noindent \textbf{Input.}
Our network takes as input a fixed-size point cloud, where 40K points are randomly sub-sampled from the original scan. In addition to using color (RGB) and coordinates (XYZ) as input features, following~\cite{qi2019votenet}, we include surface normal and a height feature of each point.

\par \noindent \textbf{Augmentation.}
We augment the input point cloud at two places in our framework: (1) data augmentation at the input, and (2) to compute the consistency loss in~\cref{eqn:consis-seg} and~\cref{eqn:consis-det}. In practice, we find it beneficial to apply different geometric transformations for the above two purposes. To augment the input, we follow~\cite{qi2019votenet} and use random sub-sampling of 40K points, random flipping in both horizontal and vertical directions, and random rotation of $[-5,5]$ degrees around the upright-axis. To compute the consistency loss, we use random flipping, point jittering, random rotation with an angle uniform in $[0,30]$ degrees around the upright-axis, random scaling by a factor from $[0.8, 1.2]$, and point dropout ($p=0.1$). Finally, we also find that jittering the point cloud is crucial to obtain good proposals for noisy  point clouds (analyzed in~\cref{sec:analysis}).

\par \noindent \textbf{Network architecture.}
\textit{(1) Backbone.} We use PointNet++~\cite{qi2017pointnet++} as the backbone model to compute the point cloud features. The model has 4 set abstraction (SA) layers and 2 feature propagation (FP) layers. The four SA layers sub-sample the point cloud to 2048, 1024, 512 and 256 points using a receptive radius of 0.2, 0.4, 0.8 and 1.2 meters respectively. The two FP layers up-sample the last SA layer's output back from 256 to 1024 points. The final output has (256+3) dimensions (feature + 3D coordinates).
\textit{(2) Segmentation module.} This module is implemented as two FP layers which upsample the backbone features (1024 points) to the input size (40K points), and a two layer MLP (implemented as two $1\times 1$ convolutional layers) which convert the features into per-point classification logits. \textit{(3) Detection module.} This module has 3 fully-connected layers, computing the classification $\bS_\text{cls}$, objectness $\bS_\text{obj}$, and final classification logits $\bS_\text{det}$ respectively, as described in~\cref{sec:det}.

\par \noindent \textbf{Training.}
We train the entire network end-to-end from scratch with an Adam optimizer for 200 epochs. We use 8 GPUs with a batch size of 32. The initial leaning rate is 0.003 and is decayed by 10$\times$ at epoch \{120, 160, 180\}.

\par \noindent \textbf{Inference.}
\textit{(1) Segmentation.} We generate the segmentation mask from the predicted logits ($\bS_\text{seg}$) by taking the class with highest score for each point. We then post-process the output for smoothness by using the detected planes (as in~\cref{eqn:smo}), and assign each point in the plane to the  most frequently occurring class. \textit{(2) Detection.} Following~\cite{qi2019votenet}, we post-process the final output probability, $\text{softmax}(\bS_\text{det})$, by thresholding to drop predictions with score $<0.01$, and class-wise non-maximum suppression (NMS) with IoU threshold 0.25.

\begin{table}[t]
    \setlength{\tabcolsep}{3pt}
    \renewcommand{\arraystretch}{1.05}
    \centering
    %\resizebox{\textwidth}{!}{
    \begin{tabular}{c| c H | c}
    \specialrule{.15em}{.05em}{.05em}
    Method & Split & axis-aligned & mIoU \\
    \hline
    \rowcolor{highlightRowColor} \multicolumn{4}{c}{\textit{Weakly-supervised methods}} \\
    \hline
    PCAM~\cite{wei2020multi} & train & \checkmark & 22.1 \\
    MPRM~\cite{wei2020multi} & train & \checkmark & 24.4 \\
    \NAME{} & train & \checkmark & {\bf 30.7} \\
    \hline
    MIL-seg & val & \checkmark& 20.7 \\
    \NAME{} & val & \checkmark& 29.6 \\
    \NAME{}+prior & val & \checkmark & {\bf 31.1} \\
    \hline
    \NAME{} & test & -$^\ast$ & 24.0 \\
    \hline
    \rowcolor{highlightRowColor} \multicolumn{4}{c}{\textit{Supervised methods}} \\
    \hline
    VoteNet~\cite{qi2019votenet} & test & - & 55.7 \\
    SparseConvNet~\cite{choy20194d} & test & - & 73.6 \\
    \specialrule{.15em}{.05em}{.05em}
    \end{tabular}%}
    \caption{
        {\bf 3D semantic segmentation on ScanNet.} \NAME{} outperforms standard baselines and existing state-of-the-art~\cite{wei2020multi}. %  by a  margin. 
    We also report fully supervised methods for reference. Note that models on train and val sets leverage axis alignment information from~\cite{qi2019votenet}, which is not present and hence not used for experiments on the test set.
    See~\cref{supp:per-cls-seg} for per-class performance.
    % \jr{($^\ast$: axis alignment matrices are missing on ScanNet test set.)}}
    \vspace{-0.15in}
    }
    \label{tab:res_seg}
    \end{table}
\begin{table}[t]
\setlength{\tabcolsep}{1pt}
\renewcommand{\arraystretch}{1.05}
\centering
\small{
%\resizebox{\columnwidth}{!}{
\begin{tabular}{c | c c c | c }
\specialrule{.15em}{.05em}{.05em}
\multirow{2}{*}{Methods} & \multicolumn{3}{c|}{Proposal} & Detection \\
    & \#boxes & MABO & AR &  mAP \\
\hline
\rowcolor{highlightRowColor} \multicolumn{5}{c}{\textit{Unsupervised methods}} \\
\hline
Qin \etal~\cite{qin2020weakly} & 1k & 0.092 & 23.6  & - \\
GSS  & $\leq$256 & 0.321 & 73.4 & - \\
GSS  & $\leq$1k & {\bf 0.378} & {\bf 86.2} & - \\
\hline
\rowcolor{highlightRowColor} \multicolumn{5}{c}{\textit{Weakly-supervised methods}} \\
\hline
MIL-det (unsup. GSS) & $\leq$1k  & 0.378 & 86.2 & 9.6 \\
\NAME{} & $\leq$1k & 0.409 & 89.3 & 18.3 \\
\NAME{}+prior & $\leq$1k & {\bf 0.427} & {\bf 90.5} & {\bf 19.7} \\
\hline
\rowcolor{highlightRowColor} \multicolumn{5}{c}{\textit{Supervised methods}} \\
\hline
F-PointNet~\cite{qi2018frustum} & - & -  & -   & 10.8 \\
GSPN~\cite{yi2019gspn}  & - & - & -  & 17.7 \\
3DSIS~\cite{hou20193d}  & - & - & - & 40.2 \\
VoteNet~\cite{qi2019votenet} & 256 & 0.436 & 84.7  & 58.6 \\
VoteNet~\cite{qi2019votenet} & 1k  & 0.450 & 88.1 & 55.3 \\
\specialrule{.15em}{.05em}{.05em}
\end{tabular}}
\caption{{\bf 3D object detection on ScanNet.} Unsupervised GSS outperforms concurrent work~\cite{qin2020weakly} by a large margin. In the weakly-supervised setting, \NAME{} outperforms standard baselines and even some fully supervised approaches~\cite{qi2018frustum,yi2019gspn}.}
\label{tab:res_det}
\end{table}

\par \noindent \textbf{Dataset.}
We use the ScanNet~\cite{dai2017scannet} and S3DIS~\cite{armeni2017joint} datasets to evaluate our method. ScanNet contains 1.2K training and 300 validation examples of hundreds of different rooms, annotated with 20 semantic categories. We extract ground truth bounding boxes from instance segmentation masks following~\cite{qi2019votenet}. To demonstrate the generalizability of our method, we futher evaluate on S3DIS, which contains 6 floors of 3 different buildings and 13 objects classes. We use the fold \#1 split following prior work~\cite{choy20194d,armeni2017joint}, where area 5 is used for testing and the rest for training.

\par \noindent \textbf{Evaluation.}
We report mean intersection over union  (mIoU) across all classes for semantic segmentation, mean average precision (mAP) across all classes  at IoU 0.25 for object detection, and average recall (AR) and mean average best overlap (MABO)  across all classes   for proposal generation. Please see~\cite{qi2019votenet, ss, dai2017scannet} for more  on these metrics.

\subsection{Baselines}
\label{sec:baseline}
Besides comparing to the few existing  3D weakly-supervised learning methods, we build the following baselines, using  standard  weakly-supervised learning techniques:
\par \noindent \textbf{MIL-seg}: Single task segmentation  trained with~\cref{eqn:mil-seg}.
\par \noindent \textbf{MIL-det}: Single task object detection, which uses the unsupervised GSS proposals and is trained with~\cref{eqn:mil-det}.
\par \noindent \textbf{\NAME{}}:  Our full model trained with~\cref{eqn:loss_seg} and~\cref{eqn:loss_det}.
\par \noindent \textbf{\NAME{}+prior}:
We compute per-class mean shapes using external synthetic datasets~\cite{shapenet2015,wu20153d}, and use those to reject proposals and pseudo labels in the \NAME{} detection module that do not satisfy the prior. We also use a floor height prior for segmentation. Please see~\cref{supp:prior} for details.

\subsection{Quantitative results on ScanNet}
\label{sec:quant}

\par \noindent \textbf{Semantic Segmentation.}
Apart from the above baselines we compare \NAME{} to recent approaches, PCAM~\cite{wei2020multi} and MPRM~\cite{wei2020multi}. PCAM can be interpreted as \textbf{MIL-seg} with a KPConv~\cite{thomas2019KPConv} backbone, and MPRM adds multiple additional self-attention modules to PCAM. Since prior work reports results on the training set only, we compare against their results on the training set in~\cref{tab:res_seg} (top 3 rows). \NAME{} outperforms both methods (PCAM and MPRM) by a significant margin (+8.6\%~/~+6.3\%). Since the main difference between prior work and our method is our joint detection-segmentation framework, these results show the effectiveness of joint-training. When comparing against our baselines on the validation set (\cref{tab:res_seg}  middle) our joint model outperforms the single-task baseline (MIL-seg) by 8.9\%. We observe a large performance gap when comparing against state-of-the-art fully supervised models (bottom two rows). One possible solution to minimize the  gap is to utilize an external object prior (\eg, shape) from readily-available synthetic data, which improves results by +1.5\%.

\begin{table}[t]
\setlength{\tabcolsep}{1pt}
\renewcommand{\arraystretch}{1.05}
%\resizebox{\columnwidth}{!}{
\centering
\small{
\begin{tabular}{c |c | c c | c }
\specialrule{.15em}{.05em}{.05em}
\multirow{2}{*}{Methods}  & Segmentation & \multicolumn{2}{c|}{Proposal} & Detection \\
 & mIoU & MABO & AR  & mAP\\
\hline
\rowcolor{highlightRowColor} \multicolumn{5}{c}{\textit{Weakly-supervised methods}} \\
\hline
MIL-seg & 17.6 & - & - & - \\
MIL-det (unsup. GSS) & - & 0.412 & 84.9 & 15.1 \\
\NAME{} & {\bf 22.3} & {\bf 0.441} & {\bf 88.3} & {\bf 19.3} \\
\hline
\rowcolor{highlightRowColor} \multicolumn{5}{c}{\textit{Supervised methods}} \\
\hline
PointNet++~\cite{qi2017pointnet} & 41.1 & - & - & -  \\
SparseConvNet~\cite{choy20194d} & 62.4 & - & - & - \\
Armeni~\etal\cite{armeni2017joint} & - & - & - & 49.9 \\
\specialrule{.15em}{.05em}{.05em}
\end{tabular}}
\caption{{\bf Generalizing to S3DIS.} \NAME{} seamlessly generalizes to S3DIS, and outperforms standard baselines for both weakly-supervised segmentation and detection.}
\label{tab:res-s3dis}
\end{table}

\begin{table}[t]
\setlength{\tabcolsep}{2pt}
\renewcommand{\arraystretch}{1.05}
\resizebox{\columnwidth}{!}{
\centering
\small{
\begin{tabular}{c|Hcccc|Hcc|cHHc}
\specialrule{.15em}{.05em}{.05em}
\multirow{2}{*}{Removed} & \multicolumn{5}{c|}{Seg.\ losses} & \multicolumn{3}{c|}{Det.\ losses}  &  Seg. & & & Det. \\
& $\mathcal{L}_\text{seg}^{\textsc{MIL}}$ & $\mathcal{L}_\text{seg}^{\textsc{SELF}}$ & $\mathcal{L}_\text{seg}^{\textsc{CST}}$ & $\mathcal{L}_{\text{d}\rightarrow \text{s}}$ & $\mathcal{L}_\text{smooth}$  & $\mathcal{L}_\text{det}^{\textsc{MIL}}$ & $\mathcal{L}_\text{det}^{\textsc{SELF}}$ & $\mathcal{L}_\text{det}^{\textsc{CST}}$ & mIoU & MABO & AR  & mAP\\
\hline
Self-training  &  \checkmark &  & \checkmark &  \checkmark & \checkmark &  & &  \checkmark &  22.1 &  0.369 & 85.4 & 13.2 \\
Cross-transformation cst. & \checkmark &  \checkmark &  &  \checkmark & \checkmark & \checkmark  & \checkmark  &  & 28.2 &  0.386 & 88.1 & 16.9   \\
Cross-task consistency & \checkmark & \checkmark & \checkmark &  & \checkmark & \checkmark & \checkmark & \checkmark  & 26.7 & 0.392 & 87.5 & 17.4 \\
Local smoothness  & \checkmark & \checkmark &  \checkmark & \checkmark &   & \checkmark &  \checkmark & \checkmark & 27.3 & 0.396 & 87.6 & 17.8 \\
\hline
\NAME{} & \checkmark & \checkmark & \checkmark & \checkmark & \checkmark & \checkmark & \checkmark &  \checkmark & {\bf 29.6} & 0.409 & 89.3 & {\bf 18.3} \\
\specialrule{.15em}{.05em}{.05em}
\end{tabular}}
}
\caption{{\bf Ablation study of losses.} We remove one set of losses at a time. All models are trained with $\mathcal{L}_\text{seg}^{\textsc{MIL}}$ and $\mathcal{L}_\text{det}^{\textsc{MIL}}$.}
\label{tab:ablation_all}
\end{table}

\par \noindent \textbf{Object Detection.}
To the best of our knowledge, no prior work has explored weakly-supervised 3D object detection using scene-level tags. We compare against our baseline methods in~\cref{tab:res_det} (middle rows). Our model significantly outperforms the single-task  baseline (MIL-det) by 8.7\% mAP, and achieves competitive results compared to even some fully supervised methods (F-PointNet~\cite{qi2018frustum} and GSPN~\cite{yi2019gspn}, numbers borrowed from~\cite{qi2019votenet}). However, the performance gap is large when compared to the state-of-the-art fully supervised methods. Similar to segmentation, the performance of our model can be further improved by incorporating an external object prior (+1.4\%).

\par \noindent \textbf{Proposal Generation.}
GSS can be made unsupervised by relying only on low-level shape and color cues, \ie, removing $s_\text{seg}$ from~\cref{eq:prop-sim} (\cref{sec:weak_prop}). We compare the unsupervised GSS to a concurrent unsupervised 3D proposal approach by Qin \etal~\cite{qin2020weakly}. We adapt their method, originally designed for outdoor environments, to indoor scenes by replacing their front-view projection to a Y-Z plane projection. For a fair comparison we use 1000 proposals and report results in~\cref{tab:res_det} (top rows). Unsupervised GSS outperforms~\cite{qin2020weakly} by a large margin, and obtains recall values comparable to even supervised approaches. The complete GSS, including the weakly-supervised similarity $s_\text{seg}$, further improves over the unsupervised baseline (+3.1\% AR/+0.031 MABO), and outperforms supervised methods on recall (+1.2\%), indicating the importance of joint training.

\subsection{Generalizing to S3DIS}
\label{sec:quant-s3dis}
We train \NAME{} on S3DIS following the settings of~\cref{sec:quant}.
Since there is no prior weakly-supervised work on this dataset, we compare against our baselines from~\cref{sec:baseline}. The results are summarized in~\cref{tab:res-s3dis}, where \NAME{} outperforms both single-task baselines with gains of 4.7\% mIoU for segmentation, 3.4\% AR for proposal generation, and 4.2\% mAP for detection. These results demonstrate that our design choices are not specific to ScanNet and  generalize to other 3D datasets.

\begin{figure}[t]
\begin{subfigure}{0.495\linewidth}
\centering
\includegraphics[height=1.1in]{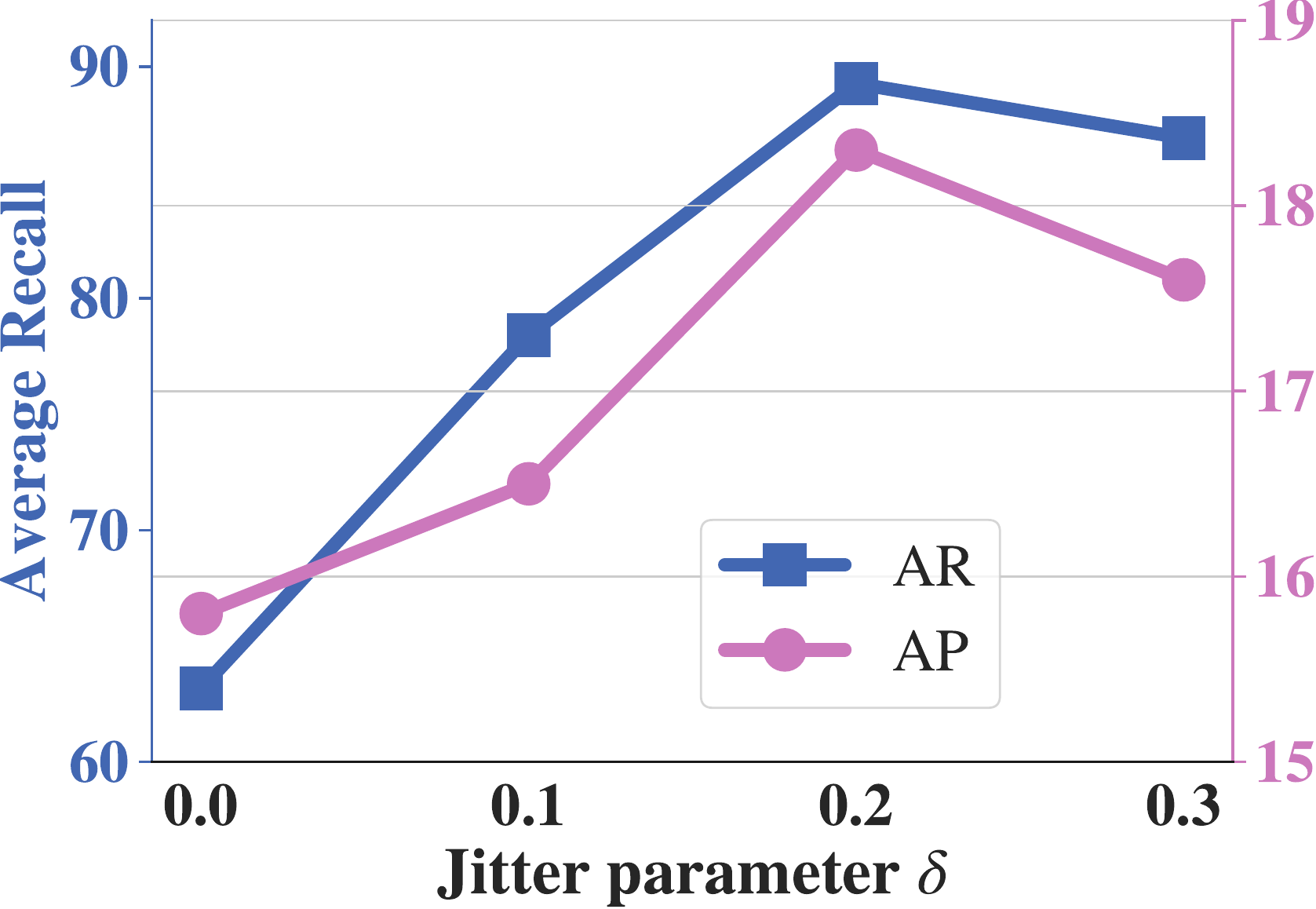}
\end{subfigure}
\hfill
\begin{subfigure}{0.495\linewidth}
\centering
\includegraphics[height=1.1in]{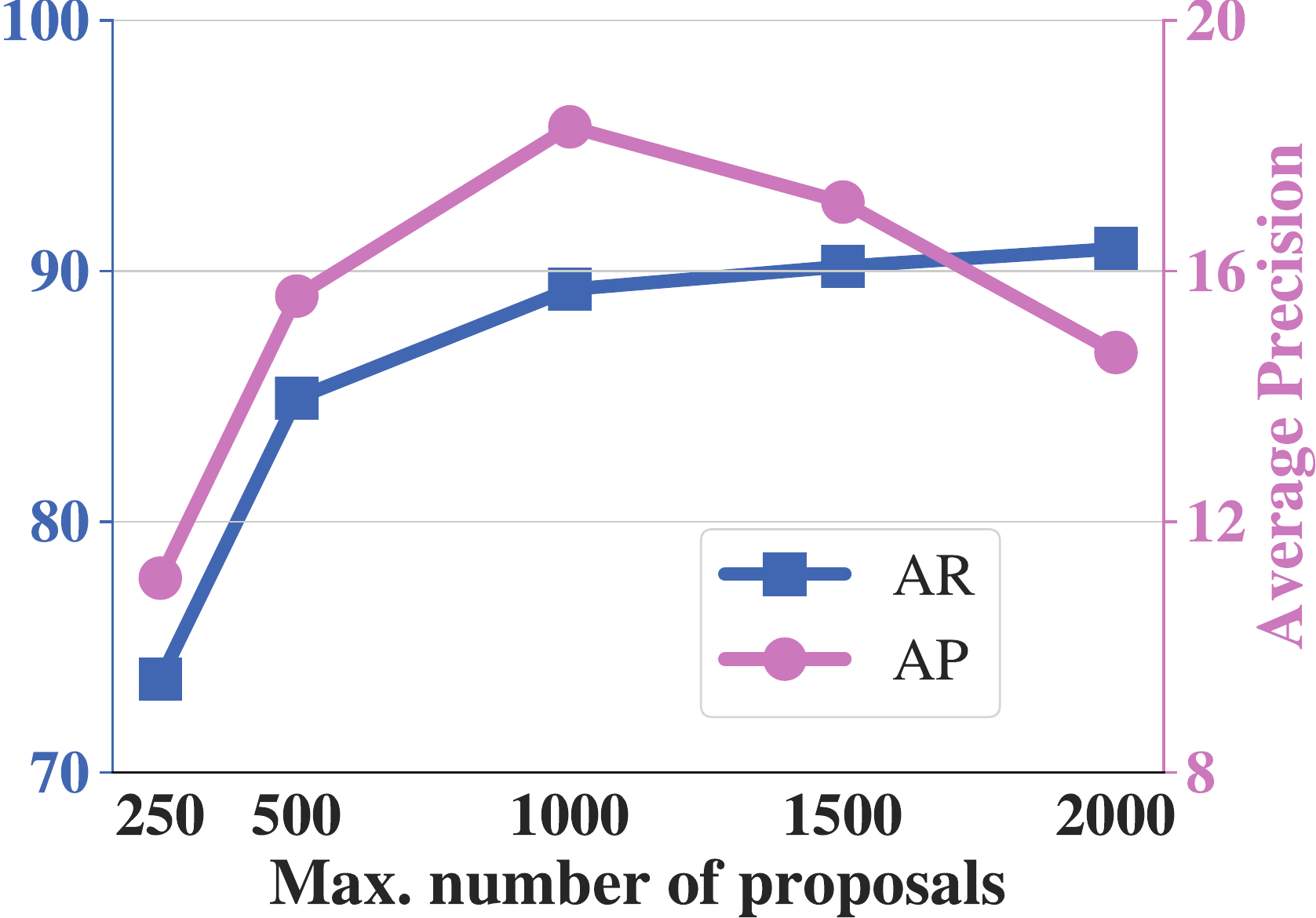}
\end{subfigure}
\caption{{\bf Effect of jittering and \#proposals.} Jittering the point cloud before proposal generation results in a $\!>\!2\%$ gain in AP. The performance varies gracefully with \#proposals, and we find 1000 proposals to have the right balance for high precision and recall.}
\label{fig:abla-prop-params} \label{fig:abla-prop}
\end{figure}

\subsection{Analysis}
\label{sec:analysis}
\par \noindent \textbf{Which loss terms matter?}
In~\cref{tab:ablation_all} we analyze the relative contribution of the loss terms in~\cref{eqn:loss_seg} and (\ref{eqn:loss_det}). We find self-training to be the most critical. Removing $\mathcal{L}_\text{seg}^{\textsc{SELF}}$ and $\mathcal{L}_\text{det}^{\textsc{SELF}}$ leads to a significant drop in both metrics: -7.5\% mIoU and -5.1\% mAP. This is consistent with observations in prior work on weak-supervision~\cite{wei2017object,ren-cvpr2018}. Next, we find enforcing consistency between detection and segmentation tasks to add large gains, especially for segmentation: 2.9\% mIoU. Enforcing consistency across transformations is particularly important for detection, leading to a 1.4\% mAP gain. Finally, encouraging  smoothness over primitive structures improves both metrics by 1.7\% mIOU and 0.5\% mAP.

\par \noindent \textbf{Jittering for proposal generation.}
We observe that scanned point clouds are often imperfect, with large holes in objects due to occlusions, clutter or sensor artifacts. This makes it challenging for GSS to correctly group parts. To overcome this, we jitter the points in 3D space using a random multiplier within range $[1-\delta/2, 1+\delta/2]$ and decide the neighboring regions based on the jittered points. This simple technique counts spatially close but non-overlapping regions as neighbors, and greatly improves GSS results. We show the impact of $\delta$ in~\cref{fig:abla-prop-params} (left).

\par \noindent \textbf{Number of proposals.}
We randomly sample at most 250, 500, 1000, 1500, 2000 regions from the same set of computed proposals and report the recall and detection mAP in~\cref{fig:abla-prop} (right). Using fewer proposals hurts both the recall and precision since the model misses many relevant objects. In contrast, a large number of proposals increases recall but hurts precision, presumably because too many proposals increase the false positive rate of the detection module. We find  $1000$ proposals to be a good balance between precision and recall, and use this number for all our experiments.

\begin{figure}[t]
\includegraphics[width=\linewidth]{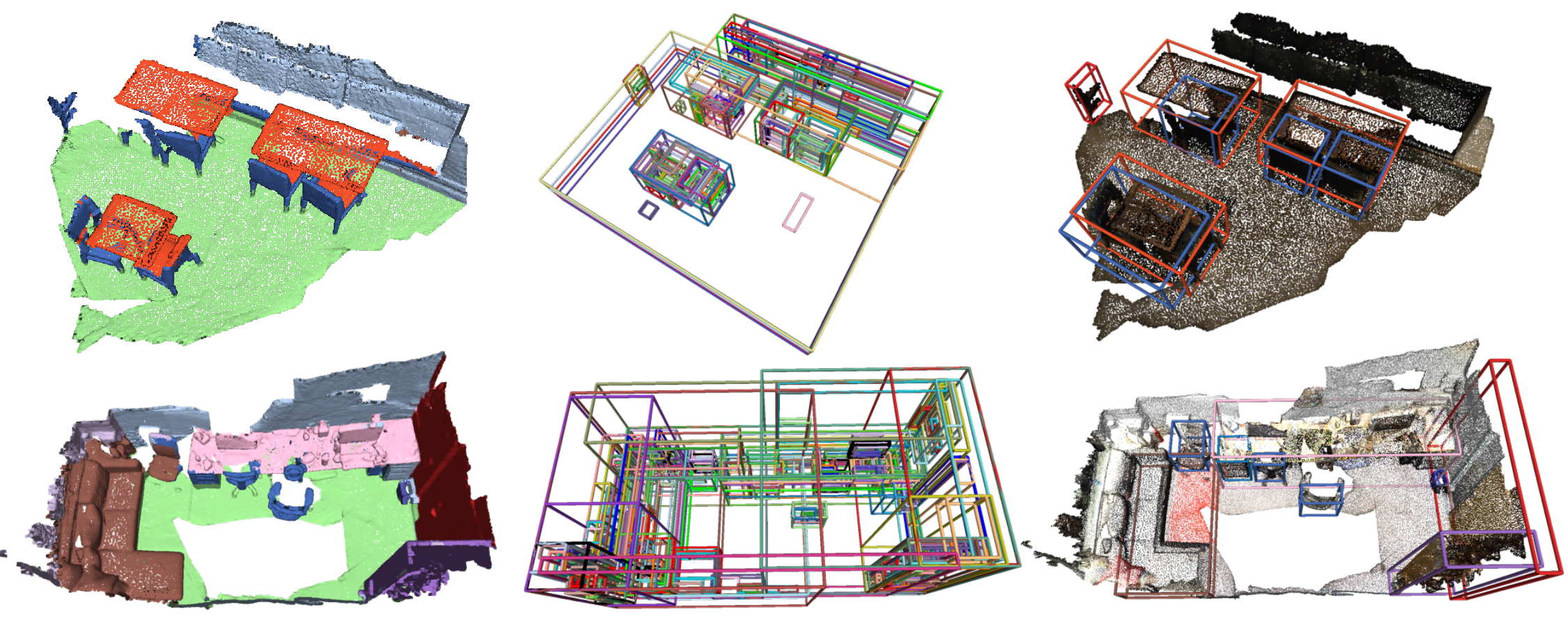}
\caption{{\bf Qualitative results on ScanNet.}
\NAME{}+prior is able to segment, generate proposals and  detect  objects  without having ever seen any spatial annotations.
\vspace{-0.1in}
}
\label{fig:quali1}
\end{figure}

\par \noindent \textbf{Qualitative results.}
\label{sec:qualitative}
\cref{fig:quali1} shows a few representative examples of our model's predictions on ScanNet. As can be seen,  input point clouds are quite challenging, with large amounts of clutter and sensor imperfections. Nevertheless, our model is able to recognize  objects such as chairs, tables, and sofa with good accuracy. Please see~\cref{supp:vis} for more results, analysis and failure modes.

%!TEX root = ../main.tex
\section{Conclusion}

We propose \NAME{}, a novel framework  for joint 3D semantic segmentation and object detection, trained using only scene-level class tags as supervision. It leverages a novel unsupervised 3D proposal generation approach (GSS) along with natural constraints between the segmentation and detection tasks. Through extensive experimentation on standard datasets we show \NAME{} outperforms single task baselines and prior state-of-the-art methods on both tasks.

\noindent\textbf{Acknowledgements.} 
This work was supported in part by NSF under Grant \#1718221, 2008387 and MRI \#1725729, NIFA award 2020-67021-32799. The authors thank Zaiwei Zhang and the Facebook AI team for helpful discussions and feedback.

\clearpage
{\small
\bibliographystyle{ieee_fullname}
\bibliography{refs}
}

%!TEX root = ../main.tex

\clearpage
\appendix
{\centering \Large \textbf{Appendix}}
\vspace{1em}

In this section, we provide: (1) algorithm details and ablation studies of Geometric Selective Search (\cref{sec:proposal}); (2) introduction of the shape detection algorithm (\cref{supp:shape}); (3) additional implementation details (\cref{supp:imp}); (4) details of the integration of an external object prior (\cref{supp:prior}); and (5) per-class segmentation results; (6) additional qualitative results (\cref{supp:vis}). 

%!TEX root = ../main.tex
\section{Geometric Selective Search (GSS)}
\label{sec:proposal}

As introduced in the main paper~\cref{sec:weak_prop}, the goal of GSS is to capture all possible object locations in 3D space. We formulate a bottom-up algorithm where the key idea is to utilize the geometric and semantic cues for guiding 3D proposal generation. 

\subsection{Approach}
Given an input point cloud with unoriented normals, we first detect primitive shapes using a region growing based method~\cite{lafarge2012creating} as detailed in \cref{supp:shape}.
It outputs a set of detected planes with assigned points, \ie, each point is assigned to at most one plane or none.

We then apply hierarchical agglomerative clustering (HAC) to generate the candidate bounding boxes from the detected planes. We first initialize a region set with the detected planes, and then compute the similarity score $s$ between all neighboring regions in the set. Two regions are neighboring if the corresponding convex hull of them overlap. To overcome the artifacts of the point cloud, we randomly jitter the points of each region before computing their convex hull. This technique greatly improves the results in practice as verified in~\cref{fig:abla-prop}. Once the neighboring relationships and similarity scores are computed, the two most similar regions are grouped into a new region. We then generate an axis-aligned 3D box for the new region as a  proposal.
New similarity scores are calculated between the resulting region and its neighbors. HAC is repeated until no neighbors can be found or only a single region remains. We provide the detailed pseudo-code in Alg.~\ref{alg:proposal}.

In order to pick which two regions $\bn_i, \bn_j$ to group, we use the similarity score  $s(\bn_i, \bn_j) = $
\begin{align} \label{eq:prop-sim}
\begin{split}
& w_1 s_\text{color}(\bn_i, \bn_j) \!+\! w_2 s_\text{size}(\bn_i, \bn_j) + \\
& w_3 s_\text{volume}(\bn_i, \bn_j) \!+\! w_4 s_\text{fill}(\bn_i, \bn_j)  \!+\!   w_5 s_\text{seg}(\bn_i, \bn_j),
\end{split}
\end{align}
where $w_i \in \{0,1\} ~~\forall i\in \{1,\cdots,5\}$ are binary indicators. Binary weights are used over continuous values to encourage more diverse outputs following~\cite{ss}. $s_\text{color}\in[0,1]$ measures the color similarity; $s_\text{size}$ and $s_\text{volume}\in[0,1]$ measure size and volume compatibility and encourage small regions to merge early; $s_\text{fill}\in[0,1]$ measures how well two regions are aligned; and $s_\text{seg}(\bn_i, \bn_j)\in[0,1]$ measures high-level semantic similarities. We detail each metric next. 

\begin{algorithm}[t]
\footnotesize
\caption{Geometric Selective Search (GSS)}
\begin{algorithmic}[1]
\Require{point cloud $\cP$}
\Ensure{3D proposal set $\cR$}
\State Detect shapes from $\cP\rightarrow$  initial regions $\cN=\{\bn_1, \bn_2, \cdots\} $
\State Initialize similarity set $\cS = \emptyset$, proposal set $\cR = \emptyset$
\For{each neighboring region pair $(\bn_i,\bn_j)$}
\State $\cS  = \cS  \cup s(\bn_i,\bn_j)$ \algorithmiccomment{compute and store similarities}
\EndFor
\While{$\cS\neq\emptyset$ } \algorithmiccomment{HAC}
\State Get the most similar pair $s(\bn_i,\bn_j)=\text{max}(\cS)$
\State Remove similarities regarding $\bn_i:\; \cS = \cS \setminus s(\bn_i,\ast)$
\State Remove similarities regarding $\bn_j:\; \cS = \cS \setminus s(\bn_j,\ast)$
\State Update region set $\cN = \cN \setminus \bn_i, \cN = \cN \setminus\bn_j$ 
\State Merge and generate new region $\bn_k = \bn_i\cup\bn_j$
\State Compute similarity of $\bn_k$ and its neighbors in $\cN$: $\cS = \cS \cup \{s(\bn_k,\bn'): \; \text{neighbor}(\bn_k,\bn')=\text{True} \;\forall \bn' \in \cN \}$
\State Add new region to $\cN = \cN \cup \bn_k$ 
\State Generate 3D proposal $\cR=\cR\cup \text{AxisAlignedBox}(\bn_k)$
\EndWhile
\end{algorithmic}
\label{alg:proposal}
\end{algorithm}

\begin{table*}[t]
\setlength{\tabcolsep}{1pt}
\centering
\resizebox{\textwidth}{!}{
\begin{tabular}{c| c c c c c c c c c c c c c c c c c c | c}
\specialrule{.15em}{.05em}{.05em}
Class & cabinet & bed & chair & sofa & table & door & window & shelf & picture & counter & desk & curtain & fridge & sc$^*$ & toilet & sink & bathtub & other & mean \\
\hline
\rowcolor{highlightRowColor} \multicolumn{20}{c}{\textit{Unsupervised GSS}} \\
\hline
ABO & 0.402 & 0.414 & 0.419 & 0.462 & 0.432 & 0.327 & 0.349 & 0.469 & 0.121 & 0.286 & 0.365 & 0.342 & 0.469 & 0.421 & 0.415 & 0.355 & 0.325 & 0.432 & 0.378 \\
Recall & 86.0 & 97.5 & 90.4 & 99.0 & 91.1 & 67.0 & 86.9 & 100.0 & 26.1 & 75.0 & 92.1 & 91.0 & 98.2 & 96.4 & 94.8 & 91.8 & 77.4 & 90.9 & 86.2 \\
\hline
\rowcolor{highlightRowColor} \multicolumn{20}{c}{\textit{GSS}} \\
\hline
ABO    & 0.449  & 0.471 & 0.441 & 0.437 & 0.464 & 0.379 & 0.388 & 0.446 & 0.136 & 0.366 & 0.381 & 0.399 & 0.501 & 0.478 & 0.409 & 0.365 &0.400 & 0.453 & 0.409 \\
Recall & 90.6 & 98.8 & 91.7 & 98.9 & 93.7 & 75.2 & 89.7 & 100.0 & 27.9 & 88.5 & 94.5 & 97.0 & 96.5 & 100.0 & 94.8 & 92.9 & 83.8 & 92.3  & 89.3  \\
\specialrule{.15em}{.05em}{.05em}
\end{tabular}}
\caption{{\bf Per-class results of GSS proposals.} GSS achieves more than 80\% recall rate for all classes except  picture (27.9\%) and door (75.2\%), where the plan detection algorithm often fails to differentiate these two objects from the surrounding wall. 
Here sc$^*$ refers to the `shower curtain' class.}
\label{tab:per_cls_prop}
\end{table*}

\par\noindent\textbf{Color similarity $s_\text{color}$.}
Color is an informative low-level cue to guide the plane grouping process. 
For each region, we first compute the L1-normalized color histogram following~\cite{ss}. The similarity score is computed as the histogram intersection:
\begin{align}
s_\text{color}(\bn_i, \bn_j)=\sum_k \min(b_i^k, b_j^k),
\end{align}
where $b_i^k, b_j^k$ are the $k$-th bin in the color histograms of $\bn_i$ and $\bn_j$ respectively. Following~\cite{ss}, we use 25 bins for each HSV color channel and 75 in total for one histogram.

\par\noindent\textbf{Size similarity $s_\text{size}$ and volume similarity $s_\text{volume}$.}
These two metrics encourage small regions to merge early. This strategy is desirable as it guarantees a bottom-up grouping of parts of different objects at multiple locations in 3D space. It encourages diverse 3D proposals and  prevents a single region from absorbing all other regions gradually.
We compute size similarity
\begin{align}
s_\text{size}(\bn_i, \bn_j)= 1 - \frac{\text{size}(\bn_i)+ \text{size}(\bn_i)}{\text{size}(\cP)}, 
\end{align}
where $\text{size}(\bn_i), \text{size}(\bn_j), \text{size}(\cP)$ are the size of the axis-aligned bounding boxes of region $\bn_i, \bn_j$ and the whole point cloud.
Similarly, volume similarity is defined as:
\begin{align}
s_\text{volume}(\bn_i, \bn_j)= 1 - \frac{\text{volume}(\bn_i)+ \text{volume}(\bn_i)}{\text{volume}(\cP)}, 
\end{align}
where $\text{volume}(\bn_i), \text{volume}(\bn_j), \text{volume}(\cP)$ are the volume of the water-tight convex hull of region $\bn_i, \bn_j$ and the whole point cloud.

\par\noindent\textbf{Alignment score $s_\text{fill}$.} 
This score measures how well two regions fit into each other and encourage merged regions to be cohesive. Essentially, if one region is contained in the other one, they should be merged first to avoid any holes. Meanwhile, a low score means the two regions don't fit very well, and they may form an unnatural region. We compute the alignment score:
\begin{align}
s_\text{fill}(\bn_i, \bn_j)= 1 - \frac{\text{size}(\bn_i\cup\bn_j) - \text{size}(\bn_i) - \text{size}(\bn_i)}{\text{size}(\cP)}, 
\end{align}
where $\bn_i\cup\bn_j$ means the union of two regions, and the other numbers are identical to the ones used for the computation of $s_\text{color}$.

\par\noindent\textbf{Semantic similarity $s_\text{seg}$.}
The above four metrics are mainly low-level geometric cues. GSS can also utilize high-level semantic information,~\ie, weakly-supervised segmentation prediction.
For each region, we first infer the segmentation mask from $\bS_\text{seg}$ using the inference procedure described in~\cref{sec:implement}. 
We then take the most likely class assignment for each point in the region and compute an L1-normalized histogram over classes for that region. The similarity score is computed as the histogram intersection:
\begin{align}
s_\text{seg}(\bn_i, \bn_j)=\sum_{c=1}^{C} \min(b_i^c, b_j^c),
\end{align}
where $b_i^c, b_j^c$ are the bin of class c in the class histograms.

\par\noindent\textbf{Post-processing.}
To remove the redundant proposals, we use several post-processing steps: (1) the proposals are first filtered by a 3D NMS module with an IoU threshold of $0.75$; (2) we then remove the largest bounding boxes after NMS as it covers the whole scene rather than certain objects due to the bottom-up nature of HAC; (3) we keep at most 1000 proposals through random sampling. 

\par\noindent\textbf{Diversification strategies.}
Since a single strategy usually overfits, we adopt multiple strategies to encourage a diverse set of proposals, which will eventually lead to a better coverage of all objects in 3D space. Specifically, we first create a set of complementary strategies, and ensemble their results afterwards. Highly-overlapping redundant proposals are removed though an NMS with IoU threshold of $0.75$ and we still keep at most 1000 proposals through random sampling after ensembeling.

\begin{table}[t]
\setlength{\tabcolsep}{1pt}
\centering
%\resizebox{\columnwidth}{!}{
\begin{tabular}{c |c |c c }
\specialrule{.15em}{.05em}{.05em}
Metric & Avg. \# boxes & MABO & AR \\
\hline
\rowcolor{highlightRowColor} \multicolumn{4}{c}{\textit{Single run}} \\
\hline
SZ & 382.9 & 0.351 & 84.1 \\
C  & 252.0 & 0.316 & 70.7 \\
V  & 366.8 & 0.367 & 84.4 \\
F  & 330.2 & 0.398 & 81.8 \\
SG & 350.7 & 0.362 & 83.9 \\
SZ+V & 373.4 & 0.366 & 84.5 \\
SZ+SG &369.2 & 0.353 & 85.1 \\
V+F  & 373.3 & 0.384 & 85.7 \\
V+SG & 385.5 & 0.362 & 83.8\\
SZ+V+SG & 377.5 & 0.391 & 86.4 \\
V+F+SG  & 381.6 & 0.380 & 84.9 \\
SZ+V+F+SG & 369.1 & 0.387 & 86.1 \\
\hline
\rowcolor{highlightRowColor} \multicolumn{4}{c}{\textit{Ensembeling}} \\
\hline
V+F, SZ+V & 712.0 & 0.378 & 86.2 \\
V+F, SZ+V+SG & 742.9 & 0.409 & 89.3 \\
\hline
\specialrule{.15em}{.05em}{.05em}
\end{tabular}
\caption{{\bf GSS results using various similarity metrics.} SZ, C, V, F, and SG represent $s_\text{size}, s_\text{color}, s_\text{volume}, s_\text{fill}$, and $s_\text{seg}$ respectively. }
\label{tab:abla_prop}
\end{table}

\subsection{Experiments}
In this sub-section we evaluate the proposal quality of GSS and validate the corresponding design choices. We evaluate on the ScanNet validation set and report the two popular metrics: average recall (AR) and mean average best overlap (MABO) across all classes. In addition, we also report the average number of boxes of each scene. 

We first examine each similarity metric and their combinations in~\cref{tab:abla_prop}. We first evaluate each single similarity and report their results in the top 5 rows, where we find size, volume, and segmentation metric to work much better than color and fill similarity. \cref{tab:abla_prop} also reports the results of different combined metrics. Combining multiple similarity metrics often yields better results than using each single similarity. The best result is achieved using the combination of size, volume, and segmentation similarities.

In practice, we find that ensembling the results of multiple runs using different similarity metrics  further improves the results as shown in \cref{tab:abla_prop} bottom. We provide the results of an unsupervised version (V+F, SZ+V) and the complete version (V+F, SZ+V+SG). Comparing these two methods, we find that introducing segmentation similarity is beneficial.

Lastly, we show per-class average best overlap (ABO) and recall rate in~\cref{tab:per_cls_prop}. We find that GSS achieves high recall rate ($>80\%$ ) for all classes except  picture (27.9\%) and door (75.2\%). This is likely due to the fact that these two objects are often embedded in the wall and hard to differentiate.

\subsection{Qualitative results}
\cref{fig:prop} illustrates several representative examples of the generated proposals on ScanNet. From left to right, we show the input point cloud, the detected shapes, GSS computed proposals, and the ground-truth boxes. We show all the  GSS computed proposals in the top 3 rows where we observe that the computed proposals are mainly around each object in the scene. In the bottom four rows, we show the best overlapping proposals with ground-truth bounding boxes. GSS generates proposals with great recall, and generalizes well to various object classes and complex scenes.

\section{Shape detection}
\label{supp:shape}
In this paper, we detect geometric shapes for two reasons: to be used in the local smoothness loss for segmentation (\cref{eqn:smo}), and as  input to the GSS algorithm (\cref{sec:proposal}). As introduced in the main paper~\cref{sec:weak_prop}, we adopt a region-growing algorithm~\cite{lafarge2012creating,cgal:ovja-pssd-20b} for detecting primitive shapes (\eg, planes). The basic idea is to iteratively detect shapes by growing regions from seed points. Specifically, we first choose a seed point and find its neighbors in the point cloud. These neighbors are added to the region if they satisfy the region requirements (\eg, on the same plane), and hence the region grows. We then repeat the procedure for all the points in the region until no neighbor points meet the requirements. In the latter case we start a new region. Region-growing  out-performs the popular RANSAC-based methods~\cite{schnabel2007efficient} because 1) it is deterministic; 2) it performs better in the presence of large scenes with fine-grained details; 3) it has higher shape detection recall. Even though it runs slower, we use it as a pre-processing step which won't influence the training speed. 

In practice, we use the efficient implementation of The Computational Geometry Algorithms Library (CGAL)~\cite{cgal:ovja-pssd-20b}. We set the search space to be the 12 nearest neighbors, the maximum distance from the furthest point to a plane to be 12, the maximum accepted angle between  a  point's normal and the normal of a plane to be 20 degree, and the minimum region size to be 50 points. We refer the reader to CGAL documents~\cite{cgal:ovja-pssd-20b} for more details. 

\begin{figure*}[t]
\includegraphics[width=0.97\linewidth]{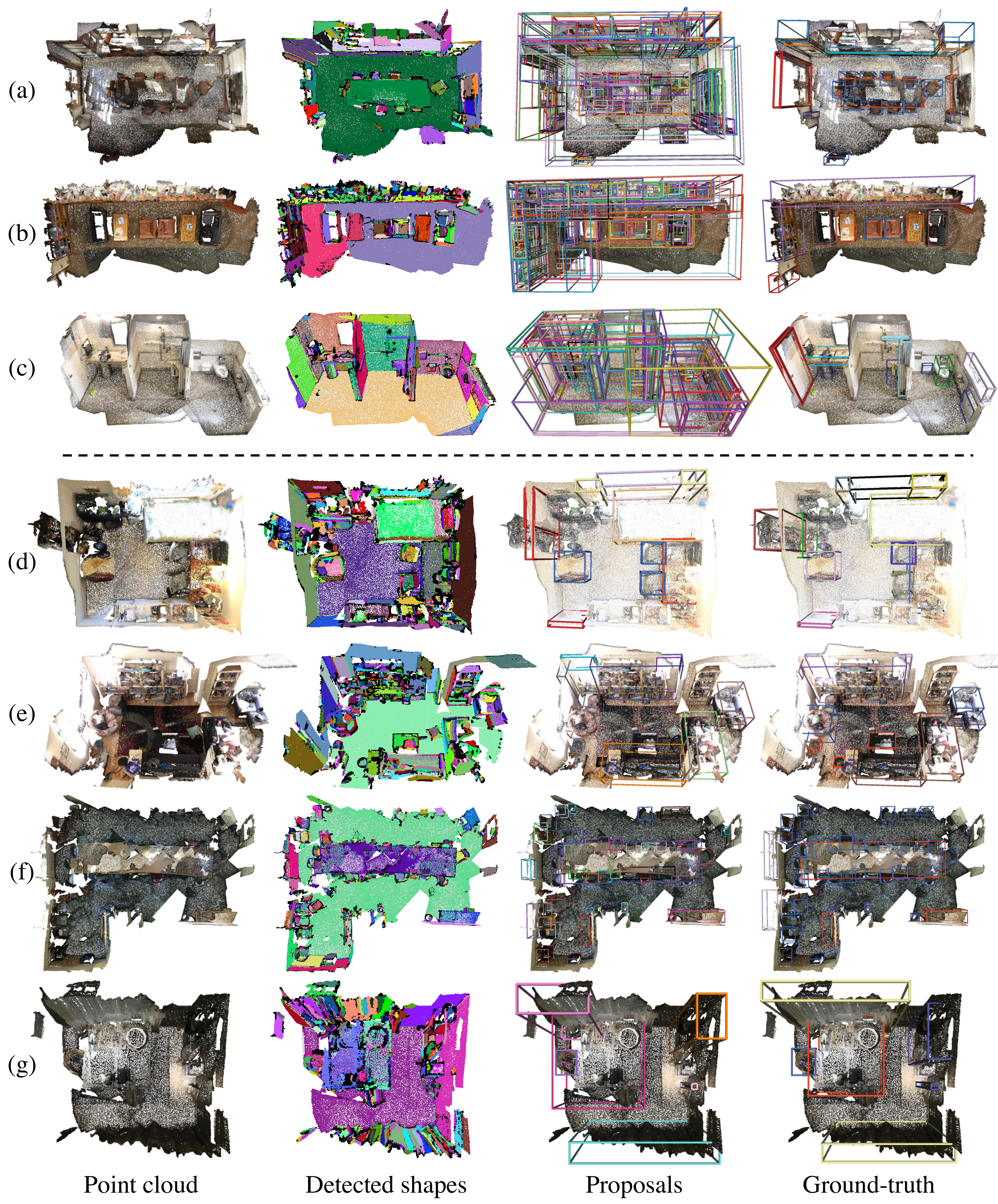}
\caption{{\bf Visualization of the computed proposals.} Top three rows show all the computed 3D proposals, from which we observe that the proposals are mainly around object areas. The bottom four rows show the proposals which best overlap with ground-truth boxes. GSS generates 3D proposals with great recall for various objects in complex scenes.}
\label{fig:supp_prop}
\end{figure*}

Representative visualization of the detected planes are provided in \cref{fig:supp_prop} second column from left. The algorithm detects big planes (\eg, floor, table top, wall) with great accuracy and doesn't over segment these regions into small pieces. This is particularly useful for WyPR as the local smoothness loss will enforce the segmentation module to predict consistently within these shapes. For complex objects (\eg, curtain, chair, and bookshelf), this algorithm segments the object regions into small shapes. Such primitive shapes will be used during the proposal generation algorithm GSS to infer the 3D bounding boxes of all objects in the scene. 

\section{Additional implementation details}
\label{supp:imp}
In this section, we provide additional implementation details.
\subsection{Geometric transformations}
\label{supp:aug}
We apply geometric transformations in two places: 1) as data-augmentation; 2) for computing cross-transformation consistency losses (\cref{eqn:consis-seg} and \cref{eqn:consis-det}) for both tasks. 

To augment the input, we first randomly sub-sample 40,000 points as input in each training iteration. We then randomly flip the points in both horizontal and vertical directions with probability 0.5, and randomly rotate them around the upright-axis with $[-5,5]$ degree. Note that after data augmentation, we only get one point cloud $\cP$ as input. 

To compute the consistency losses, we further transform the input point cloud using random  flipping of both horizontal and vertical directions with probability 0.5, larger random rotation of $[0, 30]$ degrees around the upright-axis, random scaling by a factor within $[0.8, 1.2]$, and point dropout ($p = 0.1$). We denote the resulting point cloud as $\tilde{\cP}$, which will be used when computing $\mathcal{L}_\text{seg}^{\textsc{CST}}$ and $\mathcal{L}_\text{det}^{\textsc{CST}}$.

\subsection{Backbone}
We adopt a PointNet++ network as backbone, which has four set abstraction (SA) layers and two feature propagation (FP) layers. For a fair comparison we use the same backbone network   as Qi~\etal~\cite{qi2019votenet}. The input to the backbone is a fix-sized point cloud where we randomly sample 40,000 points from the original scans. The outputs of the backbone network are geometric representations of 1024 points with dimension 3+256 (XYZ+feature dimension).

\begin{table*}[t]
\setlength{\tabcolsep}{3pt}
\centering
%\resizebox{\textwidth}{!}{
\begin{tabular}{c| c c c c c c c c c c c c c }
\specialrule{.15em}{.05em}{.05em}
metric & cabinet & bed & chair & sofa & table & door  & shelf & desk & curtain & fridge & toilet & sink & bathtub \\
\hline
$\mu_{l:w}$ & 4.64 & 1.58 & 1.29 & 1.94 & 1.65 & 5.74 & 3.17 & 1.92 & 5.78 & 1.68  & 1.55 & 1.29 & 1.93 \\
$\sigma_{l:w}$ &  5.81 & 0.45 & 0.53 & 0.54 & 1.02 & 3.78 & 2.07 &  0.91 & 3.58 & 1.16 & 0.39 & 0.26 & 0.42 \\
$\mu_{l:h}$ &  1.49 & 2.12 & 1.16 & 2.36 & 3.04 & 0.61 & 1.22 & 2.28 & 1.40 & 0.65 & 1.08 & 2.14 & 3.18 \\
$\sigma_{l:h}$ &  1.01 & 0.95 & 0.98 & 0.57 & 3.72 & 0.69 & 1.11 & 1.65 & 1.34 & 0.19 & 0.56 & 0.89 & 1.67 \\
\specialrule{.15em}{.05em}{.05em}
\end{tabular} 
\caption{{\bf Prior statistics for each class.}}
\label{tab:prior}
\end{table*}

\subsection{Segmentation module}
The segmentation module contains two feature propagation (FP) layers which upsample the geometric representations of 1024 points to 2048 and then 40,000 points with the same dimension (3+256) as before. We then use a two-layer MLP with dimension $[256, C]$ as the classifier where $C$ represents the number of classes. The segmentation module outputs a dense semantic prediction for each point in the point cloud.

\subsection{Detection module}
The detection module first applies a RoI pooling by average-pooling the features of all points  within each RoI. The computed RoI features are then fed into three fully-connected layers to get the classification $\bS_\text{cls}$, objectness $\bS_\text{obj}$, and final classification logits $\bS_\text{det}$ respectively. 

\subsection{Losses}
For computing the smoothness regularization $\mathcal{L}_\text{smooth}$  in \cref{eqn:smo}, enumerating all the detected planes in each training iteration is time-consuming and not necessary. We thus randomly sample 10 planes in each iteration, as we find 10 to be the sweet spot balancing training speed and performance. For computing the self-training losses $\mathcal{L}_\text{seg}^{\textsc{SELF}}$ and $\mathcal{L}_\text{det}^{\textsc{SELF}}$, we set the threshold $p_1$ in~\cref{alg:pseudo-seg} to be $0.1$, and $p_2$ in~\cref{alg:pseudo-det} to be $0.15$. The threshold $\tau$ in~\cref{alg:pseudo-det} is set to $0.25$.

\begin{table*}[t]
    \setlength{\tabcolsep}{1pt}
    \renewcommand{\arraystretch}{1.05}
    \centering
    \resizebox{\textwidth}{!}{
    \begin{tabular}{c| c | c c c c c c c c c c c c c c c c c c c c | c}
    \specialrule{.15em}{.05em}{.05em}
    Methods & eval. & wall & floor & cabinet & bed & chair & sofa & table & door & window & shelf & picture & counter & desk & curtain & fridge & sc$^*$ & toilet & sink & bathtub & other & mIoU \\
    \hline
    PCAM~\cite{wei2020multi} & train & 54.9 & 48.3 & 14.1 & 34.7 & 32.9 & 45.3 & 26.1 & 0.6 & 3.3 & 46.5 & 0.6 & 6.0 & 7.4 & 26.9 & 0.0 & 6.1 & 22.3 & 8.2 & 52.0 & 6.1 & 22.1 \\
    MPRM~\cite{wei2020multi} & train & 47.3 & 41.1 & 10.4 & 43.2 & 25.2 & 43.1 & 21.5 & 9.8 & 12.3 & 45.0 & 9.0 & 13.9 & 21.1 & 40.9 & 1.8 & 29.4 & 14.3 & 9.2 & 39.9 & 10.0 & 24.4 \\
    %MIL-seg & train & 39.8	& 40.1	& 14.9	& 40.1	& 26.7	& 35.6	& 11.4	& 21.8	& 23.2	& 35.3	& 10.3	& 23.1	& 23.9	& 36.2	& 9.3	& 15.1	& 8.3	& 16.2	& 23.1	& 9.7	& 23.2\\
    \NAME{} & train &  {\bf 59.3} & 31.5 & 6.4 &  {\bf 58.3} & 31.6 &  {\bf 47.5} & 18.3 &  {\bf 17.9} &  {\bf 36.7} & 34.1 & 6.2 &  {\bf 36.1} &  {\bf 24.3} &  {\bf 67.2} &  {\bf 8.7} &  {\bf 38.0} & 17.9 &  {\bf 28.9} & 35.9 & 8.2 & {\bf 30.7} \\
    \hline
    %MPRM*~\cite{wei2020multi} & val & \\
    MIL-seg & val & 36.4  & 36.1  &  {\bf 13.5}  & 37.9  & 25.1  & 31.4  & 9.6  & 18.3  & 19.8 & 33.1  &  {\bf 7.9}  & 20.3 & 21.7 & 32.5 & 6.4 & 14.0 & 7.9 & 14.7 & 19.4 & 8.5 & 20.7 \\
    \NAME{} & val &   {\bf 58.1} & 33.9 & 5.6 &  {\bf 56.6} & 29.1 &  {\bf 45.5} & 19.3 & 15.2 &  {\bf 34.2} &  {\bf 33.7} & 6.8 &  {\bf 33.3} & 22.1 & 65.6 & 6.6 &  {\bf 36.3} & 18.6 & 24.5 &  {\bf 39.8} & 6.6 & 29.6 \\
    \NAME{}+prior & val & 52.0 &  {\bf 77.1} & 6.6 & 54.3 &  {\bf 35.2} & 40.9 &  {\bf 29.6} & 9.3 & 28.7 & 33.3 & 4.8 & 26.6 &  {\bf 27.9} &  {\bf 69.4} &  {\bf 8.1} & 27.9 &  {\bf 24.1} &  {\bf 25.4} & 32.3 &  {\bf 8.7} & {\bf 31.1} \\
    \specialrule{.15em}{.05em}{.05em}
    \end{tabular}}
    \caption{{\bf 3D semantic segmentation on ScanNet.} \NAME{} outperforms standard baselines and existing state-of-the-art~\cite{wei2020multi} by a  margin. Here sc$^*$ refers to the `shower curtain' class.
    }
    \label{tab:res_cls_seg}
    \end{table*}

\section{External prior}
\label{supp:prior}
WyPR can be further improved by integrating external object priors as shown in~\cref{tab:res_seg} and~\cref{tab:res_det}. We mainly introduce two types of priors as they can be easily computed from external synthetic datasets~\cite{shapenet2015,wu20153d}: the shape prior and the location prior. 

For the shape prior, we compute the mean aspect ratio between an object's 3D bounding box length to height ($\mu_{l:h}^c$), and length to width ($\mu_{l:w}^c$) for class $c\in\{1,\cdots, C\}$. Since objects can be of arbitrary pose in 3D space, we set length and width to measure the longer and shorter edge in the XY plane. We also compute the corresponding standard deviations $\sigma_{l:h}^c$ and $\sigma_{l:d}^c$. To use it, we reject proposals whose aspect ratios don't fall within the range $[\mu_{l:h}^c - 2\sigma_{l:h}^c, \mu_{l:h}^c + 2\sigma_{l:h}^c ]$ and $[\mu_{l:w}^c - 2\sigma_{l:w}^c, \mu_{l:w}^c + 2 \sigma_{l:w}^c ]$ for any class $c\in\{1,\cdots, C\}$. We also reject pseudo bounding boxes of ground-truth class $c$ ($R^*[c]$ in~\cref{alg:pseudo-det}) whose aspect ratios don't fall in $[\mu_{l:h}^c - 2\sigma_{l:h}^c, \mu_{l:h}^c + 2\sigma_{l:h}^c ]$ and $[\mu_{l:w}^c - 2\sigma_{l:w}^c, \mu_{l:w}^c + 2 \sigma_{l:w}^c ]$. The computed statistics of each class are shown in~\cref{tab:prior}. There are certain classes that are missing from the external synthetic datasets~\cite{shapenet2015,wu20153d} such as shower curtain, window, counter, and picture. For these classes, we use the prior of other objects with similar shapes as a replacement. For example, we use the prior of curtain for shower curtain, table for counter, door for window and picture.

The location prior is only applied to the floor class. This prior is of vital importance as floor appears in almost every scene. It becomes a hard class for semantic segmentation as the MIL loss rarely sees any negative examples. Besides, a great portion of points in each scene  belongs to the floor. We estimate the floor height as the 1\% percentile of all points' heights following Qi \etal~\cite{qi2019votenet}. We force all the points below floor height to be floor. All the points above this height cannot be floor.

\section{Per-class segmentation results}
\label{supp:per-cls-seg}
In \cref{tab:res_cls_seg}, we report the per-class IoU on ScanNet. These results are consistent with \cref{tab:res_seg} in main paper. Compared to prior methods PCAM~\cite{wei2020multi} and MPRM~\cite{wei2020multi}, \NAME{} significantly out-performs them, and greatly improves the performance of some hard classes such as door, counter, and fridge.

\section{Additional qualitative results}
\label{supp:vis}
\begin{figure*}[t]
\vspace{-1em}
\includegraphics[width=0.97\linewidth]{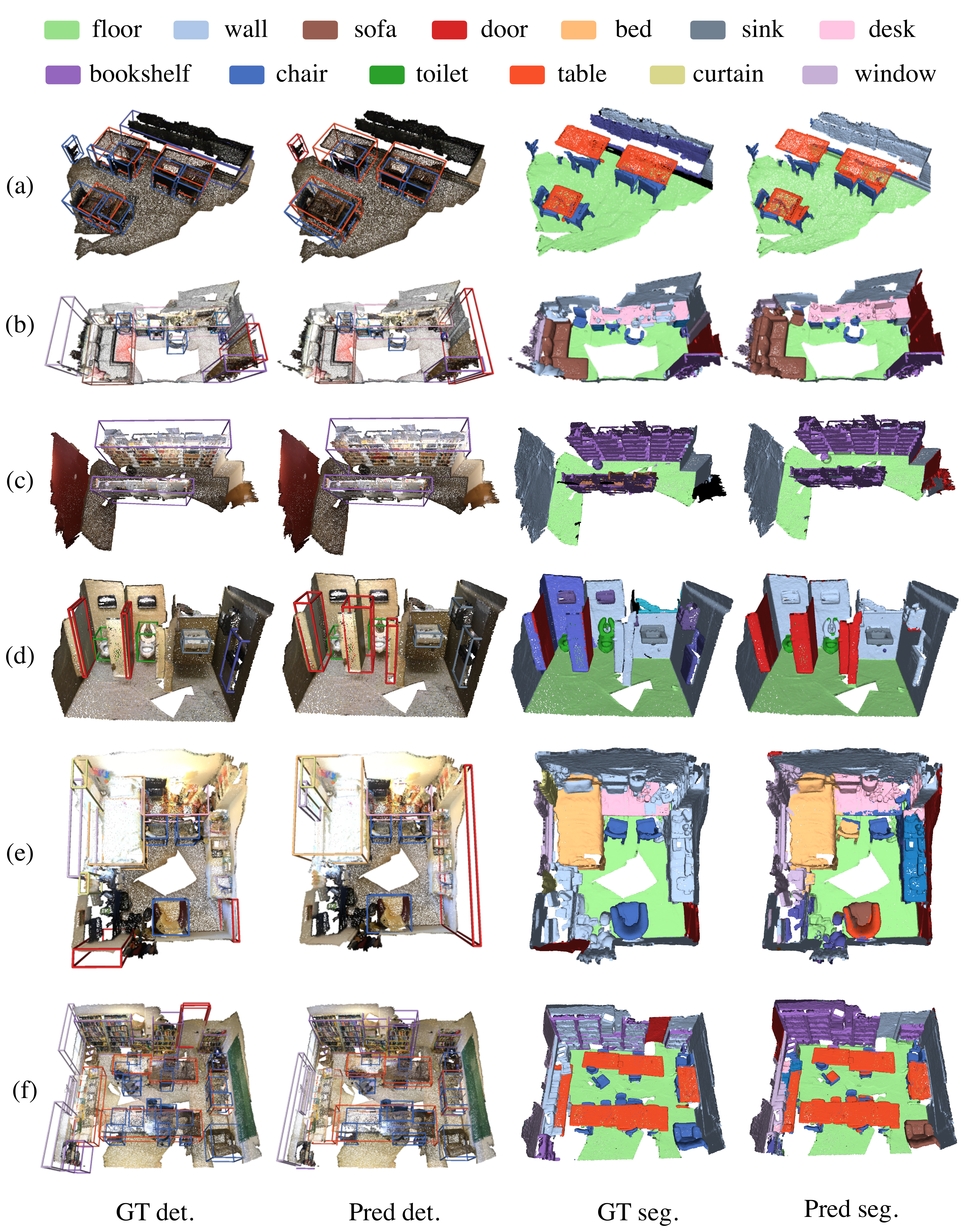}
\caption{{\bf Additional qualitative results.} We show the qualitative comparison between ground-truth labels and our (\NAME{}+prior) predictions. We show both detection and segmentation results for the same scene.}
\label{fig:supp_quali}
\end{figure*}

In~\cref{fig:supp_quali}, we show the qualitative comparison between ground-truth labels and our (WyPR+prior) prediction. In each row we show the results of both tasks for one scene. 
We find that WyPR segments and detects certain classes ({\color{table}table} in row (a, f), {\color{chair}chair} in rows (a, b, f), {\color{sofa}sofa} in row (b), {\color{bookshelf}bookshelf} in row (c, f)) with great accuracy. WyPR also learns to recognize some uncommon objects of the dataset such as {\color{green}toilet} and {\color{gray}sink} in row (d).
Moreover, we observe that predicted segmentation mask and bounding boxes are highly consistent, which reflects the effectiveness of the joint-training framework. 

Common failure cases for WyPR are partially observed objects (row (b): the window on the left side), ambiguous objects (row (a): picture and wall; row (b, f): sofa and leftmost chair). When multiple objects of the same classes are spatially close, WyPR often cannot differentiate them and only predicts one big boxes covering everything (row (a): tow chair on the left side).

\end{document}